\documentclass{article}
\PassOptionsToPackage{numbers, compress}{natbib}
\usepackage[preprint]{neurips_2023}
\usepackage{microtype}
\usepackage{enumitem}

\usepackage{eccvabbrv}
\usepackage{graphicx}
\usepackage{booktabs}
\usepackage{xcolor}         % colors
\usepackage[utf8]{inputenc} % allow utf-8 input
\usepackage[T1]{fontenc}    % use 8-bit T1 fonts

\usepackage{inconsolata}

\usepackage[accsupp]{axessibility}  % Improves PDF readability for those with disabilities.

\usepackage{hyperref}

\setcitestyle{square,numbers}
\hypersetup{colorlinks,linkcolor={blue},citecolor={green},urlcolor={magenta}}  

\usepackage[capitalize]{cleveref}
\crefname{section}{Sec.}{Secs.}
\Crefname{section}{Section}{Sections}
\Crefname{table}{Table}{Tables}

\crefname{table}{Table}{Tables}

\usepackage{url}            % simple URL typesetting

\usepackage{caption}
\usepackage{subcaption}
\usepackage{multirow}
\usepackage{tabularx}

\def\METHOD{\textsc{CRG}}
\def\METHODlong{\textsc{Contrastive Region Guidance}}
\definecolor{ggg}{gray}{0.65}

\def\llava{LLaVA}
\def\vipllava{ViP-LLaVA}
\def\gpt4v{GPT-4V}
\def\llavasmall{LLaVA-1.5-13B}
\def\llavalarge{LLaVA-1.6-34B}
\def\vipllavamodel{ViP-LLaVA-13B}
\def\rec{\textsc{Rec}}
\def\math{\textsc{Math}}
\def\ocr{\textsc{OCR}}
\def\know{\textsc{Know}}
\def\rel{\textsc{Rel}}
\def\lang{\textsc{Lang}}
\def\whatsup{What'sUp}
\def\sugarcrepe{SugarCrepe}
\def\swapatt{\textsc{Swap-Att}}
\def\swapobj{\textsc{swap-Obj}}
\def\seetrue{SeeTRUE}

\begin{document}

\title{Contrastive Region Guidance: Improving Grounding in Vision-Language Models without Training}

\author{%
  David Wan 
  \qquad
  Jaemin Cho 
  \qquad
  Elias Stengel-Eskin 
  \qquad
  Mohit Bansal \\
  UNC Chapel Hill \\
  \texttt{\{davidwan, jmincho, esteng, mbansal\}@cs.unc.edu} \\ \\
  \url{https://contrastive-region-guidance.github.io/}
}

\maketitle

\begin{abstract}
  Highlighting particularly relevant regions of an image can improve the performance of vision-language models (VLMs) on various vision-language (VL) tasks by guiding the model to attend more closely to these regions of interest.
  For example, VLMs can be given a ``visual prompt'', where visual markers such as bounding boxes delineate key image regions; this approach has become popular due to the improvement it provides in tasks requiring region-level information.  
  However, current VLMs that can incorporate visual guidance are either proprietary and expensive or require costly training on curated data that includes visual prompts.  
  We introduce \textbf{\METHODlong{}} (\METHOD{}),
  a training-free guidance method that enables open-source VLMs to respond to visual prompts. 
  \METHOD{} contrasts model outputs produced with and without visual prompts, 
  factoring out biases revealed by the model when answering \emph{without} the information required to produce a correct answer (\ie{}, the model's prior).
  \METHOD{} achieves substantial improvements in a wide variety of VL tasks: When region annotations are provided, \METHOD{} increases absolute accuracy by up to $11.1\%$ on ViP-Bench, a collection of six diverse region-based tasks such as recognition, math, and object relationship reasoning.
  We also show \METHOD{}'s applicability to spatial reasoning, where we obtain up to $10\%$ improvement on the hardest setting of \whatsup{}, as well as to compositional generalization -- improving accuracy by $11.5\%$ and $7.5\%$ on two challenging splits from \sugarcrepe{} -- and to image-text alignment for generated images, where we improve by up to $8.4$ AUROC and $6.8$ F1 points on \seetrue{}.
  For cases that do not have reference regions for the prompt, we also show that \METHOD{} allows us to re-rank regions proposed by an object detection model in referring expression comprehension and phrase grounding benchmarks like RefCOCO/RefCOCO+/RefCOCOg and Flickr30K Entities, with an average improvement of $3.2\%$ in accuracy when multiple proposals are available. In our analysis, we explore alternative masking strategies for \METHOD{},
  demonstrate how \METHOD{} impacts the model's probability over relevant text phrases, and evaluate the role of the region guidance strength, empirically validating \METHOD{}'s design choices. 
\end{abstract}

\section{Introduction}
\label{sec:intro}

Recent progress in large vision-language models (VLMs) has led to significant advances in tackling multimodal tasks by marrying the language-based reasoning strength of large language models (LLMs) with a visual encoder such as ViT \cite{dosovitskiy2020image}. 
While large VLMs (\eg{}, \llava{} \cite{liu2024visual, liu2024llavanext}, BLIP \cite{li2023blip}, PaLI \cite{chen2023pali}, \etc{}) have increasingly strong performance on tasks involving a whole image (\eg{}, answering questions about images \cite{antol2015vqa, goyal2017making} or describing them \cite{young2014image, kazemzadeh2014referitgame}), 
they often struggle with grounding specific regions,
making errors on inter-object spatial relations~\cite{kamath2023whats} and compositional reasoning~\cite{hsieh2023sugarcrepe}.
This inability to ground also prevents models from following ``visual prompts'' \cite{focalclick,kirillov2023segany,zou2023segment,Shtedritski2023,Cai2023VIP-LLAVA,Yang2023SoM}, where visual markers like bounding boxes are overlaid onto the image to help the model focus on important regions.
Improving models' visual prompt following ability has the potential to increase performance across a wide variety of VL domains where fine-grained reasoning is key, including visual question answering, 
image-text alignment,
spatial reasoning, and referring expression comprehension.

\begin{figure}[t!]
    \centering
    \includegraphics[width=0.95\textwidth]{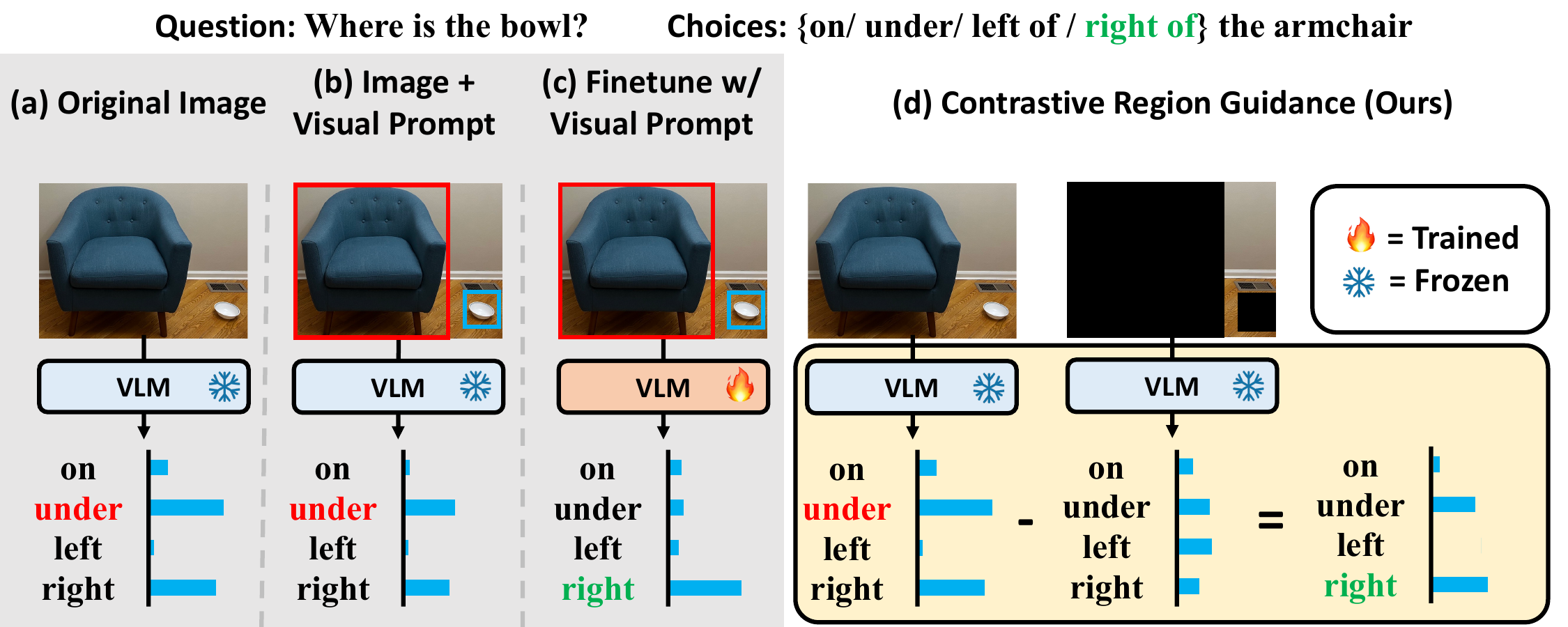}
    \caption{
    Comparison of different methods for visual grounding. (a) Predicting the answer with a base VLM fails. (b) Even when bounding boxes are added, open-source VLMs produce the wrong answer. (c) The VLM can be trained to recognize overlays like bounding boxes, but this process involves updating the VLM and is costly.
    (d) Our method, \METHOD{}, offers a way to correct predictions without training. The right image has relevant object regions blacked out. Here, the model's distribution reflects its prior on answering \emph{``under''} and \emph{``left''} even without visual evidence.
    By factoring this distribution out, we reduce the prior, leading to the correct answer.
    }
    \vspace{-1em}
   \label{fig:teaser}
\end{figure}

For example, in \cref{fig:teaser} (a), the base VLM struggles with a question that requires spatial reasoning, \emph{``Where is the bowl?''},
mistakenly answering that the bowl is under the chair (while the bowl is to the right of the chair). 
The failure can in part be attributed to the model's \emph{prior}, which biases the output towards certain answers even in the absence of relevant information; for example, in \cref{fig:teaser} (d), we see that even when the objects are blacked out, the model still tends to answer \emph{``under''}, despite the fact that the question is unanswerable from the masked image, as the region under the chair is masked.  

Several approaches for correcting these errors and improving fine-grained region grounding have been attempted, but require costly, proprietary models or additional data and training. 
Yang~\etal{}~\cite{Yang2023SoM} introduce Set-of-Mark (SoM) prompting, a method that overlays visual markers directly onto an image at test time,
helping the model to generate answers that are grounded to specific image regions.
However, SoM is tested only on \gpt4v{}, and our results in \cref{tab:vip_bench} and \cref{tab:whatsup_ablations}
indicate that SoM with segmentation marks does not transfer well to open-source VLMs.
Furthermore, as illustrated in \cref{fig:teaser} (b), when the model is given an image overlaid with bounding boxes as marks, it predicts probabilities for the question that are similar to those predicted when using the original model depicted in \cref{fig:teaser} (a).
SoM's reliance on \gpt4v{} leads to a number of limitations: firstly, the model used is financially costly and large, making it impractical for many applications. In fact, the authors only present results on a small subset of the data
due to ``limited quota and absence of \gpt4v{} API''.
Secondly, the model's training data and details are unknown, meaning that it may in fact have been finetuned using additional data to supervise grounding. 
This kind of finetuning has been shown to improve open-source VLMs' ability to follow visual prompts: Cai~\etal{}~\cite{Cai2023VIP-LLAVA} synthesize large amounts of fine-tuning data for adding visual markings like arrows and bounding boxes to images to enable open-source VLMs to follow visual prompts.
While finetuning is effective, as illustrated in \cref{fig:teaser} (c) where the fine-tuned model correctly predicts the correct preposition with high confidence, it incurs a substantial training cost, especially as models grow in size.
To address the shortcomings of existing methods (\ie{}, reliance on costly training or proprietary models), we propose a \emph{training-free} method that is compatible with a variety of existing models.
We additionally show that our method is complementary to models finetuned with region grounding supervision, \ie{}, it can further increase a model's performance when using visual prompts. 

Specifically, we propose \textbf{\METHODlong{}} (\METHOD{}), a novel strategy that leverages classifier-free guidance (CFG) \cite{ho2021classifierfree,sanchez2023stay} to help open-source VLMs focus on specific regions and understand visual markers \emph{without additional training}.
\METHOD{} reduces a given model's 
bias towards certain answers (\eg{}, towards \emph{``under''} in \cref{fig:teaser}) by factoring out
what the model's response would be without visual evidence from the key region.
Intuitively, after factorization, the final answer will be the one that changed the most when the key visual information is removed (\ie{}, the answer that relies most heavily on the visual information), whereas all answers that do not rely on visual information from the key region will be down-weighted. 
As shown in \cref{fig:teaser} (d), by blacking out the relevant objects, \METHOD{} reveals a prior that biases the model towards the incorrect answer, \emph{``under''}; in other words, the model answers \emph{``under''} even in the absence of the relevant visual evidence it would need to determine the relationship between objects.
\METHOD{} then factors this prior out, amending the answer distribution, and providing the correct answer, \emph{``right''}.
Crucially, \METHOD{} relies only on either visual prompts or -- if such prompts are not provided -- access to an object detection module for proposing bounding boxes; such modules are readily available across many domains \cite{zhang2022glipv2, liu2023grounding}.

We evaluate \METHOD{} on a variety of datasets in 5 different domains and on 2 different models, described in more detail below.
\begin{itemize}[leftmargin=10pt]
    \item \textbf{Visual Prompt Following.} To measure \METHOD{}'s ability to incorporate visual prompts, we test on ViP-Bench \cite{Cai2023VIP-LLAVA}, which contains 6 diverse task types each requiring understanding granular, region-level reasoning: Object Recognition (\rec{}), \ocr{}, commonsense knowledge (\know{}), \math{}, relations (\rel{}), and language generation (\lang{}). 
    For example, the \math{} subset of ViP-Bench requires solving math equations based on images of multiple equations, where one is highlighted or circled, like in \cref{fig:method_figure} (a). 
    Here, \METHOD{} improves on average $11.1\%$ accuracy over \llavalarge{} model, performing competitively with the strongest baseline of \gpt4v{}.
    When applied to \vipllava{}, \METHOD{} also provides substantial improvements, indicating it is complementary to supervised methods. 
    \item \textbf{Spatial Reasoning.} We also measure the role \METHOD{} plays in improving spatial reasoning by highlighting relevant image regions (see \cref{fig:teaser}); on the hardest setting of the \whatsup{} spatial reasoning benchmark \cite{kamath2023whats}, \llavasmall{} with \METHOD{}  outperforms the baselines by up to $8.3\%$, and in fact also surpasses training-based methods relying on large amounts of pre-training with the same model by $15.4\%$.
    Furthermore, \llavalarge{} with \METHOD{} improves by $10\%$ over a \llavalarge{} baseline on \whatsup{}'s hardest setting. 
    \item \textbf{Compositional Generalization.} Furthermore, we show that \METHOD{}'s better grounding leads to improvements in visual understanding and reasoning.
    We find that \METHOD{} helps to address a major limitation of current vision-language methods: a poor ability to analyze language compositionally.
    Models often cannot differentiate between two similar sentences like ``a plant on a house'' and ``a house on a plant''~\cite{thrush2022winoground, ma2023crepe, ray2024cola}.
    We show that with \METHOD{}, \llavalarge{} improves its performance on \sugarcrepe{} \cite{hsieh2023sugarcrepe} -- a challenging benchmark dataset for compositionality in visual-language tasks -- by $11.5\%$ and $7.5\%$ over the model without \METHOD{} and by $4.7\%$ and $3.6\%$ with \llavalarge{} over the strongest \gpt4v{} baseline on \sugarcrepe{}'s two challenging settings. 
    \item \textbf{Evaluation on Images from Text-to-Image Generation Models.} We show that \METHOD{} can also evaluate generated images; when applied to \citet{yarom2023what}'s DrawBench, EditBench, and COCO-t2i splits, \METHOD{} improves a model's ability to identify matching image-text pairs by $8.4$ AUROC and $6.7$  F1 points on average.
    \item \textbf{Reranking for Referring Expression Comprehensions and Phrase Grounding.} Because of its granularity, \METHOD{} can be used to rerank bounding box proposals from an object detector to find ones relevant to a given text (see \cref{fig:method_figure} (d) for an example); on the RefCOCO, RefCOCO+, and RefCOCOg \cite{kazemzadeh2014referitgame,Mao2015GenerationAC} referring expression comprehension task and the Flickr30K Entities phrase grounding task \cite{flickrentitiesijcv}, \METHOD{} applied to \llavasmall{} improves performance by up to $3.2\%$ over a baseline \llavasmall{} reranker on cases with multiple bounding boxes.

\end{itemize}

We further conduct a detailed analysis of \METHOD{}, first by ablating each of its components. Our findings highlight that \METHOD{}'s masking strategy, \ie{}, blacking out each object separately, proves to be the most effective and outperforms alternative contrastive approaches that vary in granularity of the black-out regions, such as blacking out the entire image or blacking out the objects with segmentation masks.
Our analysis also reveals that current models fail to follow the prompts using other popular visual prompting strategies that do not use contrast, such as only overlaying bounding boxes and segmentation masks. 
Furthermore, we examine the impact of \METHOD{} on the probability of grounded text that is aligned with specific regions, confirming that it increases the likelihood of correct text and penalizes incorrect text. 
This underscores \METHOD{}'s precision in enhancing model interpretability.
Finally, our experiments demonstrate that the default value of the guidance strength, \ie{}, how much the model should rely on the contrast, consistently achieves high performance across different tasks, validating the robustness of our configuration.

\section{Related Work}
\label{sec:related_work}

\paragraph{Visual prompting for VLMs.}
Several recent research directions have studied prompting VLMs by manipulating visual inputs in different ways:
(i) incorporating learnable soft tokens in visual inputs for parameter-efficient finetuning~\cite{bahng2022exploring,Khattak2023Maple},
(ii) concatenating an image sequence as a demonstration of a new task~\cite{Bar2022,bai2023sequentialLVM},
and
(iii) grounding regions by overlaying visual markers (\eg{}, masks/boxes/circles, \etc{}) onto visual inputs~\cite{Yao2021CPT,Zellers2021Merlot,Shtedritski2023}.
Our work falls into to the third category, using visual guidance for grounding. 
Yang~\etal{}~\cite{Yang2023SoM} present set-of-mark (SoM) prompts, where an image is partitioned into regions with segmentation models and each region is marked with a number marker, which improves visual grounding for \gpt4v{}~\cite{openai2023gpt4}.
However, in our experiments detailed in \cref{sec:vip-bench,sec:ablation_whatsup} we confirm past findings~\cite{Cai2023VIP-LLAVA} that such visual prompt does not work well with public VLMs such as \llava{}.
Cai~\etal{}~\cite{Cai2023VIP-LLAVA} instruction-tune with diverse visual markers so that VLMs can better follow visual prompts from user input.
Instead of relying on proprietary models or finetuning,
our work elicits visual grounding in VLMs by masking image regions and contrasting model distributions, \ie{}, with no additional training or data. 
Moreover, we show that our work is complementary to finetuning methods such as those used by Cai~\etal{}~\cite{Cai2023VIP-LLAVA}, obtaining additional improvements when combined.

\paragraph{Context-guided sampling for autoregressive models.}
Several works in different domains have proposed context-guided sampling for autoregressive models
to incorporate additional context.
Guided models can be thought of as sampling tokens from the logit difference of conditional and unconditional models:
$\text{logit} (y|c,x) - \text{logit} (y|x)$
where $x$ is the input, $y$ is the output, and $c$ is the context
(see \cref{sec:method} for additional details).
For text generation,
Shi~\etal{}~\cite{shi2023trusting} extend contrastive decoding~\cite{li-etal-2023-contrastive},
by contrasting the logits of conditional and unconditional language models. 
CFG has also been applied in multimodal settings: for autoregressive image generation, Gafni~\etal{}~\cite{Gafni2022Make-A-Scene} use classifier-free guidance~\cite{ho2021classifierfree} to incorporate contextual inputs (\ie{}, text and segmentation map).
For image captioning, Kornbilith~\etal{}~\cite{Kornblith2023CFGCaption} use classifier-free guidance (CFG), contrasting the logits of an image captioner and a language model.
Concurrently, Leng~\etal{}~\cite{Leng2023HallucinationVCD}
and Zhao~\etal{}~\cite{zhao2024mitigating} use CFG to improve the faithfulness of VLMs by adding Gaussian noise to the entire image or object detection results to text input. 
While all of these existing methods combining CFG with image manipulate the entire image (via dropping~\cite{Kornblith2023CFGCaption} or adding noise~\cite{Leng2023HallucinationVCD}),
our work, \METHOD{} differs in focusing on fine-grained guidance, explicitly grounding to specific image regions, \ie{}, operating at the \emph{sub-image} level.

\paragraph{Biases and lack of grounding in visual models.}
\METHOD{}'s benefit comes from factoring out the biases present in VL models and tasks, whereby the correct response can be obtained \emph{without} considering the relevant image regions, or, in some cases, without the image entirely.
Such biases have been well documented in past work \cite{zhang2016yin, goyal2017making, cho2023generative}. 
Other work has noted that VQA models often focus on non-relevant regions of images even when correctly answering questions and has attempted to regularize models towards focusing on relevant regions \cite{selvaraju2019taking, wu2019self, liu2022answer}. 
Along these lines, \citet{ying2022visfis} introduce a series of losses for mitigating cases where models are ``right for the wrong reasons'', \ie{}, the answer is right but based on unrelated regions of the image, with some losses using human-drawn
bounding boxes. 
\METHOD{} also aims to draw attention to relevant image regions, but does so in a gradient-free manner and can operate using automatically detected bounding boxes.

\section{Method}
\label{sec:method}

\subsection{Background: Visual Prompting for VLMs
}
\label{sec:method_background}

In our setting, a vision-language model (VLM) with parameters $\theta$ takes as input an image $I \in \mathbb{R}^{H \times W \times 3}$ and a text $X = [x_1,...,x_n]$ of $n$ tokens,
and outputs a text $Y = [y_1,...,y_m]$ with $m$ tokens.
When generating the output text $Y$, we generate tokens autoregressively from a probability distribution conditioned on the input $I$ and $X$. 
At time $t$, the probability of the token $y_t$ is:
\begin{equation}
    y_t \sim p_{\theta} (y_t | I,X,y_{<t})
    \propto \exp{ \text{logit}_{\theta}(y_t | I,X,y_{<t})}
    \label{eq:prob}
\end{equation}

where $\text{logit}_{\theta}$ is the unnormalized log probability of token $y_t$, \ie{}, before softmax. 
Recent work \cite{Shtedritski2023,Yang2023SoM,Cai2023VIP-LLAVA} has introduced
\textit{visual prompting}
methods
that augment images by overlaying visual markers (\eg{}, bounding boxes, masks, and arrows) to
highlight specific regions.
While past work has found that visual prompting improves visual grounding of \gpt4v{}~\cite{Yang2023SoM} or VLMs specifically trained on images with visual prompts~\cite{Cai2023VIP-LLAVA},
we find that publicly available base VLMs usually ignore such visual prompts in our experiments (\cref{tab:vip_bench}, \ref{tab:visual_understanding}, and \ref{tab:seetrue}).

\begin{figure}[t!]
    \centering
    \includegraphics[width=\textwidth]
    {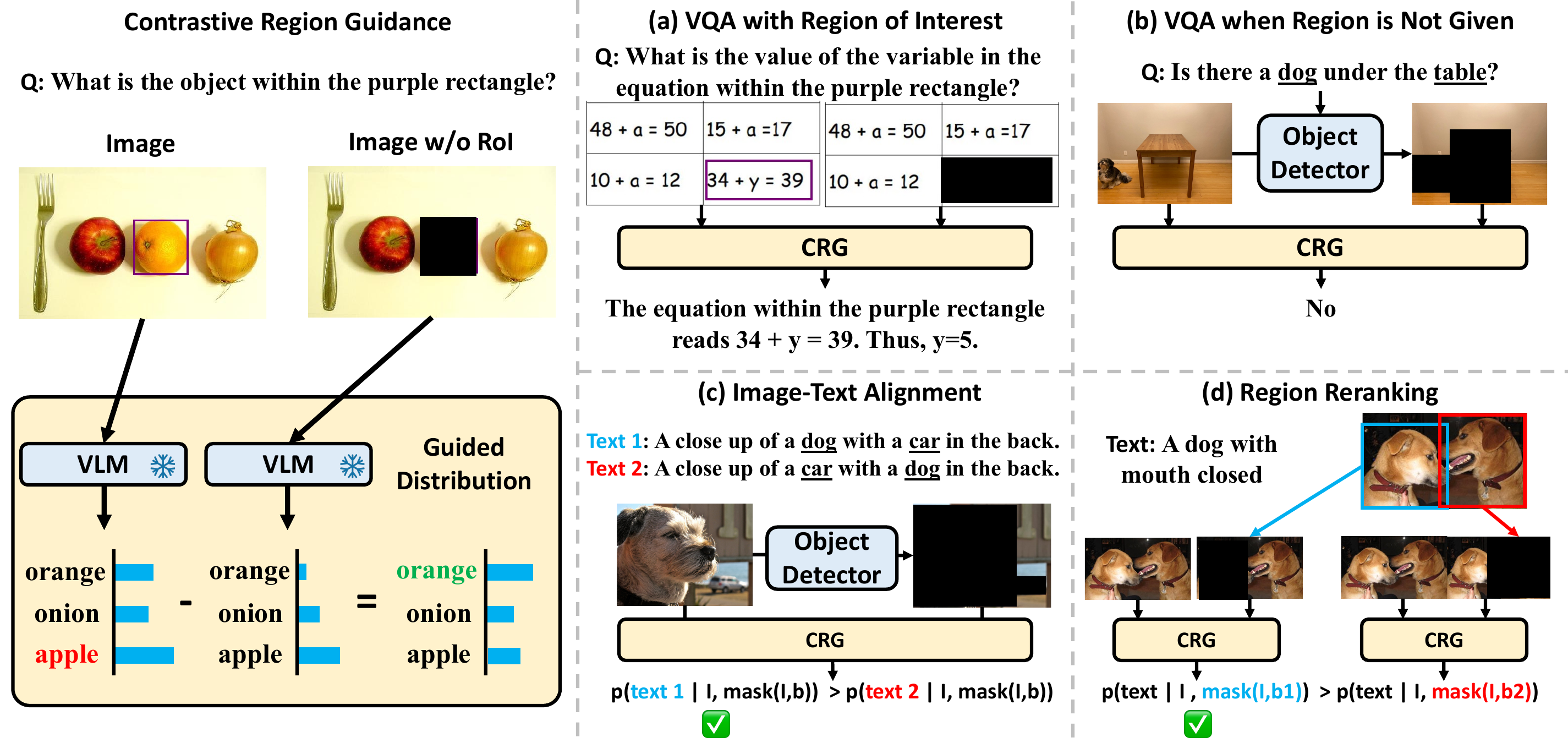}
    \caption{
    \textbf{Left}: Illustration of our method, Contrastive Region Guidance (\METHOD{}), which guides VLMs to focus on specific regions of interest (ROI).
    \textbf{Right}: Applications of \METHOD{} to various VL tasks:
    \textbf{(a)}: When answering a visual question with ROI, \METHOD{} guides a VLM to answer about the specific region.
    \textbf{(b)}: Even when no specific regions are provided, we can leverage an object detector to find important objects and guide the VLM to focus on the objects.
    \textbf{(c)}: For image-text alignment, \METHOD{} guides the model in generating text related to objects and their relationships found in images, leading to a higher probability of the correct text versus the incorrect text.
    \textbf{(d)}: \METHOD{} can also help VLMs to find the region corresponding to a given text from a set of multiple region proposals by finding the mask that provides the largest contrast. 
    }   
    \label{fig:method_figure}
\end{figure}

\subsection{Contrastive Region Guidance (CRG) for Visual Grounding in VLMs}
\label{sec:our_method}

We introduce \METHODlong{} (\METHOD{}), a training-free visual grounding method that guides any VLM to focus on specific regions in an image by extending classifier-free guidance (CFG)~\cite{ho2021classifierfree}. Inspired by work on visual feature importance~\cite{ribeiro-etal-2016-trust,ying2022visfis}, we measure the importance of an image region by how a VLM's output distribution changes when the region is removed, and use the contrast between distributions to guide a VLM to focus on a particular region, as illustrated on the left side of \cref{fig:method_figure}. Concretely, we sample outputs from a probability distribution derived by contrasting an image $I$ with another image $I'=\texttt{mask}(I, b)$, where the pixels in a specific region $b$ are masked out with black pixels:
\begin{align}
y_{t} & \propto p_{\theta} (y_t|I,X,y_{<t}) (\frac{p_{\theta}(y_t|I,X,y_{<t})}{p_{\theta}(y_{t}|\texttt{mask}(I,b),X,y_{<t})})^{\alpha}\\
& \sim \texttt{softmax}[ (1 + \alpha) \cdot \text{logit}_{\theta}(y_{t} | I,X,y_{<t}) - \alpha \cdot \text{logit}_{\theta}(y_t|\texttt{mask}(I,b),X,y_{<t}) ].
\label{eq:cfg}
\end{align}

Here, $\alpha$ is the region guidance strength parameter that controls the strength of the focus on the region $b$. A larger $\alpha$ means the region guidance is more amplified; for example, $\alpha=1$ puts a high weight on the region and $\alpha=0$ reduces the equation to standard decoding. We follow prior work~\cite{shi2023trusting,obrien2023contrastive} to use $\alpha=1$ for all settings.

As shown in \cref{fig:method_figure}, \METHOD{} is applicable to many VL tasks, including image-conditional text generation as well as image-text and region-text alignment tasks. When a region of interest is given as in \cref{fig:method_figure} (a), we can guide VLMs to focus on that region while generating answers.
When no specific region is given as in \cref{fig:method_figure} (b) and (c), we can use region proposals from text-conditional object detectors such as GroundingDINO~\cite{liu2023grounding} and guide VLMs to focus on proposed regions. We do so by taking all noun phrases (``dog'' and ``table'' in \cref{fig:method_figure} (b), and ``dog'' and ``car'' in \cref{fig:method_figure} (c)), finding their corresponding bounding boxes, and blacking out the objects in the image.
We then either generate the answers in VQA (\eg{}, \cref{fig:method_figure} (b)) or forced-decode the sentence and retrieve its probability (\eg{}, \cref{fig:method_figure} (c)). 
For cases where there are multiple bounding box candidates for the NP, we apply the following re-ranking strategy.
For each bounding box proposal, we black out the corresponding image region and calculate the score of the given text or phrase with~\cref{eq:cfg}. 
We select the region that achieves the highest contrast when blacked out. 
As illustrated in \cref{fig:method_figure} (d), removing the dog on the left causes the most drastic change in the probability of the sentence "a dog with mouth closed", thereby indicating a strong association with the described text. 

\section{Experiments and Results}
\label{sec:experiments}

We demonstrate the usefulness of \METHOD{} in diverse vision-language tasks. First, we show that \METHOD{} can unlock the visual prompt following capabilities of VLMs on ViP-Bench \cite{Cai2023VIP-LLAVA} (\cref{sec:vip-bench}).
Next, we demonstrate the effectiveness of \METHOD{} for improving image-text alignment in VLMs on three datasets (\cref{sec:img-text-alignment}): \whatsup{} \cite{kamath2023whats}
that measures spatial understanding, \sugarcrepe{} \cite{hsieh2023sugarcrepe} that measures compositional generalization, and \seetrue{} \cite{yarom2023what} whose images are from text-to-image generation models. Moreover, we show that \METHOD{} can also be used as a re-ranker for visual grounding tasks on four datasets: RefCOCO, RefCOCO+ \cite{kazemzadeh2014referitgame}, RefCOCOg \cite{Mao2015GenerationAC}, and Flickr 30K Entities \cite{flickrentitiesijcv} (\cref{sec:rec}). Lastly, we also provide three ablation studies in (\cref{sec:ablation}), comparing different methods for contrasting regions, evaluating the probability shifts for correct and incorrect texts, and analyzing the impact of region guidance strength $\alpha$.

\subsection{Experimental Setup}
\label{sec:exp_setting}

We use the \llavasmall{} \cite{liu2023visual} and \llavalarge{} \cite{liu2024llavanext} models. 
For visual prompt following with ViP-Bench, we use the provided visual prompts in the dataset; for other tasks where no region is given, we first extract noun phrases with spaCy\cite{Spacy2020},
then generate region proposals for each phrase with GroundingDINO-B \cite{liu2023grounding}, and filter the resulting bounding boxes to those with scores above a threshold of 0.3.
For text generation, we opt for greedy decoding to ensure reproducibility. 
We use \METHOD{} strength $\alpha = 1$ for all settings. We refer the readers to \cref{sec:experimental_setup_appendix} for more details on the experimental setup and dataset statistics. 

\subsection{Evaluation on Visual Prompt Following}
\label{sec:vip-bench}

ViP-Bench \cite{Cai2023VIP-LLAVA} is comprised of 303 image-question pairs specifically designed to comprehensively evaluate visual prompt following capabilities,
with six categories:
Object Recognition (\rec{}), OCR (\ocr{}), Knowledge (\know{}), Math (\math{}), Object Relationship Reasoning (\rel{}), and Language Generation (\lang{}).
We report the performance on the default split - \textit{synthesized visual prompts} - consisting of tight bounding boxes.
In addition to the baselines provided in the paper, we also apply Set-of-Mark \cite{Yang2023SoM} (SoM) approach to our models. 
Specifically, we use the reference bounding box to generate the segmentation mask with SAM \cite{kirillov2023segany}, and subsequently overlaying the mask and attach numbers to the image, as described in \cite{Yang2023SoM}. To make a fair comparison, we transform the questions asking about the bounding boxes into questions asking about the numbers, which SoM expects. Details can be found in \cref{sec:som_appendix}.

\begin{table}[tb]
    \caption{ViP-Bench \cite{Cai2023VIP-LLAVA} results. * indicates results from \cite{Cai2023VIP-LLAVA} and $\dag$ indicates models fine-tuned with visual prompt data. For each model we run, we report the average and standard deviation of 5 runs using ViP-Bench~\cite{Cai2023VIP-LLAVA}'s GPT-4-based evaluation, and \textbf{bold} the best prompting or guidance strategy. 
    }
    \label{tab:vip_bench}
    \centering
    \resizebox{\textwidth}{!}{%
    \begin{tabular}{l c c c c c c c}
    \toprule
    Model & \rec{} & \ocr{} & \know{} & \math{} & \rel{} & \lang{} & \textsc{Avg} \\
    \midrule
    Shikra 7B$\dag$* \cite{chen2023shikra} & 40.2 & 10.0 & 28.0 & 3.5 & 18.9 & 20.6 & 33.7 \\
    GPT4ROI 7B$\dag$* \cite{zhang2023gpt4roi} & 35.6 & 16.7 & 29.7 & 9.7 & 32.5 & 13.8 & 35.1 \\
    Qwen-VL-Chat* \cite{Qwen-VL} & 43.0 & 30.4 & 40.2 & 9.7 & 25.7 & 28.7 & 39.2 \\
    InstructBLIP-13B* \cite{dai2023instructblip} & 42.5 & 12.2 & 37.5 & 3.2 & 33.2 & 12.5 & 35.8 \\
    \gpt4v{}* \cite{openai2023gpt4} & 58.1 & 48.5 & 69.5 & 63.3 & 82.9 & 68.1 & 55.9 \\
    \midrule
    \llavasmall{} \cite{liu2023visual} & 45.9$_{\pm 0.4}$ & 18.7$_{\pm 0.4}$ & \textbf{46.4$_{\pm 0.9}$} & 6.5$_{\pm 0.0}$ & \textbf{50.5$_{\pm 2.7}$} & \textbf{40.5$_{\pm 1.9}$} & 40.2$_{\pm 0.4}$ \\
    + SoM \cite{Yang2023SoM} & \textbf{48.9$_{\pm 0.2}$} & 14.4$_{\pm 0.1}$ & 46.0$_{\pm 0.3}$ & 3.2$_{\pm 0.0}$ & 34.3$_{\pm 0.0}$ & 30.9$_{\pm 0.8}$ & 41.2$_{\pm 0.2}$ \\
    + \METHOD{} (Ours) & 48.0$_{\pm 0.3}$ & \textbf{20.0$_{\pm 0.5}$} & 45.1$_{\pm 1.0}$ & \textbf{10.3$_{\pm 1.4}$} & 46.1$_{\pm 2.0}$ & 	32.6$_{\pm 3.6}$ &  \textbf{41.8$_{\pm 0.2}$} \\
    \midrule
    \vipllava{}-13B$\dag$ \cite{Cai2023VIP-LLAVA} & 56.9$_{\pm 0.1}$ &  \textbf{22.7$_{\pm 1.9}$} & \textbf{54.0$_{\pm 1.1}$} & 11.7$_{\pm 1.1}$ & 49.3$_{\pm 1.6}$ & \textbf{53.1$_{\pm 4.2}$} & 48.6$_{\pm 0.2}$\\
    + SoM \cite{Yang2023SoM} & 49.3$_{\pm 0.1}$ & 19.1$_{\pm 0.1}$ & 43.8$_{\pm 0.5}$ & 5.8$_{\pm 0.0}$ & 41.1$_{\pm 0.0}$ & 35.1$_{\pm 1.9}$ & 41.8$_{\pm 0.1}$ \\
    + \METHOD{} (Ours) & \textbf{57.4$_{\pm 0.3}$} &  20.6$_{\pm 0.5}$ & \textbf{54.0$_{\pm 1.0}$} & \textbf{12.1 $_{\pm 0.2}$} & \textbf{49.4$_{\pm 1.8}$} & 45.0$_{\pm 1.3}$ & \textbf{49.4$_{\pm 0.2}$} \\
    \midrule
    \llavalarge{} \cite{liu2024llavanext} & 49.1$_{\pm 0.3}$ & 28.7$_{\pm 0.4}$ & 48.1$_{\pm 0.4}$ & 41.1$_{\pm 1.0}$ & 45.6$_{\pm 0.7}$ & \textbf{49.5$_{\pm 2.9}$} & 45.0$_{\pm 0.2}$ \\
    + SoM \cite{Yang2023SoM} & 42.0$_{\pm 0.7}$ & 41.1$_{\pm 1.3}$ & 48.5$_{\pm 1.0}$ & 39.7$_{\pm 2.7}$ & 31.1$_{\pm 2.4}$ & 47.0$_{\pm 2.5}$ & 42.0$_{\pm 0.8}$ \\
    + \METHOD{} (Ours) & \textbf{58.4$_{\pm 0.3}$} & \textbf{47.5$_{\pm 0.5}$} & \textbf{61.1$_{\pm 0.3}$} & \textbf{50.3$_{\pm 1.8}$} & \textbf{61.6$_{\pm 0.9}$} & 46.5$_{\pm 1.1}$ & \textbf{56.1$_{\pm 0.3}$} \\
    \bottomrule
    \end{tabular}
    }
\end{table}

\textbf{\METHOD{} unlocks visual prompt following, matching fine-tuned models.}
We present the result in \cref{tab:vip_bench}. We note that the base model \llavasmall{} already surpasses several baselines including fine-tuned visual prompting models like Shikra \cite{chen2023shikra} and GPT4ROI \cite{zhang2023gpt4roi}, as well as other notable VLMs such as Qwen-VL-Chat \cite{Qwen-VL} and InstructBLIP \cite{dai2023instructblip}.
Nevertheless, applying \METHOD{} on the \llavasmall{} model results in further improvements of $2.1\%$, $1.3\%$, $3.8\%$ in the \rec{}, \ocr{}, and \math{} categories, respectively, and improve $1.6\%$ on average.
While \llavasmall{} with \METHOD{} lags behind \vipllavamodel{} \cite{Cai2023VIP-LLAVA}, which uses the \llavasmall{} as backbone but is trained with curated visual prompting data, the performance gap between \llavasmall{}+\METHOD{} and \vipllava{} narrows significantly in the \ocr{} and \math{} categories. This indicates that \METHOD{} can help models follow visual prompts by contrasting between an image and a version of it where the visual prompt region is removed.

\textbf{\METHOD{} can also help models fine-tuned with visual prompts.}
Our findings also indicate that \METHOD{} is \emph{complementary} to fine-tuned models for visual prompting, \ie{}, \vipllava{},  via its contrast between distributions, further improving the performance on categories like \rec{}, \math{}, and \rel{} by $0.5\%$, $0.4\%$, and $0.1\%$, respectively, with an average improvement of $0.8\%$.

\textbf{\METHOD{} is more helpful to a stronger VLM backbone.}
The improvement is more pronounced when we apply \METHOD{} to the \llavalarge{} model, achieving on average an $11.1\%$ increase in performance.
Despite a $3\%$ decrease in \lang{}, the improvements in other categories range from $9.2\%$ to $18.8\%$, surpassing all previous models. 
Notably, \llavalarge{}+\METHOD{}
also surpasses \vipllavamodel{} in all categories except \lang{}, although it was never trained with any visual prompt data.
This underscores the efficiency of \METHOD{} in scaling up models without the need for additional training.

\textbf{Set-of-Mark prompting \cite{Yang2023SoM} is not effective on \llava{}-based models.} 
Finally, we observe that Set-of-Mark (SoM) generally decreases the performance of \llava{}-based models, indicating that this visual prompting strategy, which works on proprietary models, does not transfer well to the open-source VLMs we study. One potential reason is that SoM requires OCR capability, a domain where \llava{}-based models perform poorly compared to \gpt4v{} ($28.7\%$ for \llavalarge{} without \METHOD{} versus $48.5\%$ for \gpt4v{}).
While we observe improved overall performance when SoM is applied to \llavasmall{}, this is driven solely by increased recognition performance, which improves by $3\%$.
In all other categories, SoM decreases performance, sometimes drastically (\eg{}, a $16.2\%$ drop on \rel{}). Similarly, we observe that applying SoM to \vipllavamodel{} and \llavalarge{} decreases accuracy in all categories except \ocr{} and \know{} for \llavalarge{}, lowering performance by an average of $6.8\%$ and $3\%$ for \vipllavamodel{} and \llavalarge{}, respectively.

\subsection{Evaluation on Image-Text Alignment (Spatial Understanding, Compositionality, Generated Image Evaluation)}
\label{sec:img-text-alignment}

The region-text scores \METHOD{} produces can be used to measure the alignment between an image and a piece of text. 
We apply this to answering questions about spatial understanding by scoring answers, to compositional reasoning, where we use \METHOD{}'s score to decide between possible descriptions, and to evaluating generated images, where we score the match between an image description and a model-generated image. 

\subsubsection{Evaluation on Spatial Understanding}\label{sec:whatsup}
\whatsup{} \cite{kamath2023whats} is a benchmark for assessing the spatial understanding capabilities of VLMs, with 820 images of unambiguous spatial relations between two household objects (\eg{}, a chair and a bowl, \etc{}), where the images contain only the two objects in four distinct spatial relations (see \cref{fig:teaser} and \ref{fig:method_figure} (b)).
As illustrated in \cref{fig:method_figure} (b),
we extract bounding boxes for the two objects.
We compare our method with the best-performing baselines, including FLAVA \cite{singh2022flava}, CLIP~\cite{DBLP:conf/icml/RadfordKHRGASAM21}, and \gpt4v{}~\cite{openai2023gpt4}.

\begin{table}[tb]
    \caption{Results on spatial understanding (\whatsup{}) and compositionality (\sugarcrepe{}) benchmarks tasks. * indicates results reported in  \cite{kamath2023whats,hsieh2023sugarcrepe}.
    }
    \label{tab:visual_understanding}
    \centering
    \small
    \begin{tabular}{l ccc cc }
    \toprule
    & \multicolumn{3}{c}{\whatsup{}} & \multicolumn{2}{c}{\sugarcrepe{}} \\
    \cmidrule(lr){2-4} \cmidrule(lr){5-6}
    Model & Indiv. & Pairs & Set of 4 & \swapobj{} & \swapatt{} \\
    \midrule
    CLIP ViT-L-14 \cite{DBLP:conf/icml/RadfordKHRGASAM21} & 26.1* & 1.5* & 0.0* & 60.1* & 62.3*  \\
    CLIP RN50x64 \cite{DBLP:conf/icml/RadfordKHRGASAM21} & 26.2 & 2.0 & 0.0 & 61.8* & 66.7* \\
    FLAVA \cite{singh2022flava} &  30.4* & 10.9* & 0.0* & - & -\\
    \vipllava{}-13B \cite{Cai2023VIP-LLAVA} & 70.9 & 57.5 & 21.8 & 74.3 & 84.2  \\
    \gpt4v{} \cite{openai2023gpt4} & - & - & - & 83.1* & 90.1* \\
    \midrule
    \llavasmall{} \cite{liu2023visual} & 73.1 & 60.6 & 28.9 & 78.0 & 83.9 \\
    + bbox overlay & 71.0 & 57.1 & 22.0 & 75.1 & 83.8 \\
    + \METHOD{} (Ours) & \textbf{76.7} & \textbf{64.2} & \textbf{37.2} & \textbf{84.5} & \textbf{91.3} \\ 
    \midrule
    \llavalarge{} \cite{liu2024llavanext} & 86.8 & 76.8 &  54.0 & 76.3 & 86.2 \\
    + bbox overlay & 82.2 &  69.4 &  39.7 & 76.7 & 85.0 \\
    + \METHOD{} (Ours) & \textbf{87.6} & \textbf{80.3} & \textbf{64.0} & \textbf{87.8} & \textbf{93.7}  \\
    \bottomrule
    \end{tabular}
\end{table}

\textbf{\METHOD{} improves spatial understanding in VLMs.}
The results, as shown in \cref{tab:visual_understanding}, demonstrate that \METHOD{} consistently improves accuracy across all settings when applied to \llavasmall{}, improving accuracy by $3.6\%$ for both individual and pairs settings, and by $8.3\%$ for the set of 4 setting.
Notably, \METHOD{} with \llavasmall{} again outperforms \vipllavamodel{} -- despite the latter's extensive additional training -- when prompted with the bounding boxes. On \llavalarge{}, \METHOD{} also increases accuracy on all settings. For the hardest `Set of 4' setting, which involves accurately linking four prepositions to their corresponding images, \METHOD{} improves the accuracy by $10\%$. Interestingly, we see that applying bounding boxes as visual markers on the images (`+ bbox overlay') does not improve the performance for both models, indicating that visual prompting is not something the base VLM can already do and thus illustrating the effectiveness of \METHOD{}.

\subsubsection{Evaluation on Vision-Language Compositionality}\label{sec:sugarcrepe}
The \sugarcrepe{} \cite{hsieh2023sugarcrepe} dataset evaluates the compositional reasoning capability of VLMs, highlighting the fact that they generally struggle in correctly identifying instances when objects or attributes are swapped. Focusing on the subsets \swapobj{} and \swapatt{} -- the two subsets that current models have the most difficulty with \cite{hsieh2023sugarcrepe} -- we include best-performing models in \cite{hsieh2023sugarcrepe}, including CLIP \cite{DBLP:conf/icml/RadfordKHRGASAM21} and \gpt4v{} \cite{openai2023gpt4}.

\textbf{\METHOD{} improves compositional generalization of VLMs.}
As depicted in \cref{tab:visual_understanding}, our observations reveal a consistent pattern where \METHOD{} enhances the performance of the original models. Applying \METHOD{} to \llavasmall{} results in improvements of $6.5\%$ and $7.4\%$ in the \swapobj{} and \swapatt{} subsets, respectively, while for \llavalarge{} it improves the two tasks by $11.5\%$ and $7.5\%$. 
Notably, applying \METHOD{} to \llavasmall{} surpasses the performance of \gpt4v{} by $1.3\%$ on average, indicating \METHOD{}'s effectiveness in improving models' compositional understanding.

\subsubsection{Evaluation on Images from Text-to-Image Generation Models}
\label{sec:seetrue}

Next, we show how \METHOD{} can also be applied for text-to-image scenarios where we have generated images.
For this, we employ \seetrue{} \cite{yarom2023what}, a meta-evaluation benchmark to assess the model's ability to determine whether the given image-text pair is aligned or not.
methods. The dataset contains real text and synthetic images, encompassing examples from DrawBench \cite{saharia2022photorealistic}, EditBench \cite{10204528}, and COCO \cite{lin2014microsoft},
containing 1,311, 3,827, and 1,791 image-text pairs, respectively.
The authors collected 3 human annotations of binary judgments per example
for the three benchmarks.
We follow the authors in measuring performance with Area Under the ROC Curve (AUROC) 
and 
additionally include F1, which we compute by taking a threshold on the score and labeling instances above the threshold as positive. Since no validation or training data for the three sets are available for threshold tuning, we set the threshold to the average score the model assigns to all examples for each dataset. 

\textbf{\METHOD{} helps measure the alignments between text and generated images.}
We observe a similar trend that adding \METHOD{} improves the performance greatly, increasing AUROC by $7.3$ and F1 points by $5.4$ on average when applied to \llavasmall{} and $8.4$ AUROC and $6.7$ F1 points for the 34B model. 
\METHOD{} can also be combined with \vipllava{} resulting in an increase of $8.7$ points on AUROC and $6.2$ points on F1, again complementing the learned visual prompt-following. 
We also observe that directly visually prompting the models (`+ bbox overlay') does not improve the results, even when using the fine-tuned \vipllava{} model.
This verifies the effectiveness and robustness of \METHOD{} for evaluating model-generated images.

\begin{table}[tb]
    \centering
    \caption{Results on text-to-image evaluation benchmarks from \seetrue{}.}
    \label{tab:seetrue}
    \small
    \begin{tabular}{l cc cc cc cc}
    \toprule
    Model & \multicolumn{2}{c}{DrawBench} & \multicolumn{2}{c}{EditBench} & \multicolumn{2}{c}{COCO-t2i} & \multicolumn{2}{c}{Average of 3 datasets} \\
    \cmidrule(lr){2-3} \cmidrule(lr){4-5} \cmidrule(lr){6-7} \cmidrule(lr){8-9} 
    & AUROC & F1  & AUROC & F1  & AUROC & F1 & AUROC & F1  \\
    \midrule
    CLIP ViT-L14 \cite{DBLP:conf/icml/RadfordKHRGASAM21} & 61.4 & 51.5 &  62.1 & 60.4 & 59.3 & 61.1 & 60.9 & 57.7 \\
    CLIP RN50x64 \cite{DBLP:conf/icml/RadfordKHRGASAM21} & 60.7 & 50.8 & 67.1 & 65.3 & 58.6 & 59.6 & 62.1 & 58.6 \\
    VQ\textsuperscript{2}\cite{yarom2023what} & 70.4 & 59.6 & 60.8 & 63.3 & 67.7 & 66.1 & 66.3 & 63.0 \\
    \midrule
    \llavasmall{} \cite{liu2023visual} & 62.9 & 53.3 & 62.8 & 63.3 & 60.4 & 53.3 & 62.1 & 56.6  \\
    + bbox overlay & 63.1 & 52.5 &  63.0 & 63.2 & 60.0 & 52.5 & 62.1 & 56.1 \\
    + \METHOD{} (Ours) & \textbf{68.3} & \textbf{58.2} & \textbf{71.7} & \textbf{69.6} & \textbf{68.3} & \textbf{58.2} & \textbf{69.4} & \textbf{62.0} \\
    \midrule
    \vipllavamodel{} \cite{Cai2023VIP-LLAVA} & 60.3 & 52.2 & 63.0 & 62.5 & 60.3 & 64.1 & 61.1 & 59.6 \\
    + bbox overlay & 60.3 & 52.0 & 63.4 & 62.4 & 60.3 & 63.4 & 61.3 & 59.3 \\
    + \METHOD{} (Ours)  & \textbf{68.5} & \textbf{58.6} & \textbf{71.8} & \textbf{69.6} &  \textbf{69.2} & \textbf{69.2} & \textbf{69.8} & \textbf{65.8} \\
    \midrule
    \llavalarge{} \cite{liu2024llavanext} & 70.9 & 60.0 & 66.4 & 63.8 &  59.9 & 61.8 & 65.7 & 61.9 \\
    + bbox overlay & 70.6 & 59.9 & 66.5 & 64.1 &  59.4 & 62.3 & 65.5 & 62.0 \\
    + \METHOD{} (Ours)  & \textbf{77.6} & \textbf{63.1} & \textbf{75.7} & \textbf{72.9} & \textbf{69.0} & \textbf{69.9} & \textbf{74.1} & \textbf{68.6} \\
    \bottomrule
    \end{tabular}
    
\end{table} 

\subsection{Evaluation on Referring Expression Comprehension and Phrase Grounding}
\label{sec:rec}

Finally, we evaluate \METHOD{}'s abilities in referring expression comprehension (REC), \ie{}, locating the object referred to by a sentence, and phrase grounding, \ie{}, locating multiple objects referred to by a phrase. Specifically, we test whether the model can assign the correct bounding box given a textual description by re-ranking bounding box proposals such that the top bounding box matches a given phrase. 
We include three classic grounding benchmarks, RefCOCO, RefCOCO+ \cite{kazemzadeh2014referitgame}, and RefCOCOg \cite{Mao2015GenerationAC} for REC , and Flickr30K Entities \cite{flickrentitiesijcv} for phrase grounding.
For each proposal, we assign scores to bounding boxes based on the model's probability of producing the phrase when overlaid on the image. Following prior work \cite{liu2023grounding,li2021grounded}, we evaluate the methods using accuracy@0.5, where we consider a predicted box correct if it has an IoU greater than 0.5 with to the reference box.

\begin{table}[tb]
    \caption{
    Results on region re-ranking experiments for 
    referring expression comprehension and phrase grounding benchmarks.
    We report Top-1 accuracy@0.5. For each dataset, we include the percentage of the examples where multiple proposals are available in the parentheses.}
    \label{tab:visual_grounding}
    \centering
    \resizebox{\textwidth}{!}{%
    \begin{tabular}{l cc cc c c }
    \toprule
     & \multicolumn{2}{c}{RefCOCO} & \multicolumn{2}{c}{RefCOCO+} & RefCOCOg & Flickr30K Entities \\
     \cmidrule(lr){2-3}
     \cmidrule(lr){4-5}
     \cmidrule(lr){6-6}
     \cmidrule(lr){7-7}
    Model & testA ($23.3\%$)& testB ($26.9\%$) &  testA ($28.2\%$) & testB ($38.9\%$) & test ($27.2\%$) & test ($26.9\%$)\\
    \midrule
    \multicolumn{7}{c}{\textit{Full Dataset}}\\
    \midrule
    GroundingDINO-B &  77.3 & 72.5 & 72.0 & 59.3 & 66.3 & 70.4 \\
    + \llavasmall{} &  80.4 & 72.8 & 75.8 & \textbf{60.4} & 68.6 & 71.2  \\
    + \llavasmall{} + \METHOD{} (Ours) & \textbf{81.6} & \textbf{73.2} &  \textbf{77.0} & 60.0 & \textbf{69.6} & \textbf{72.8}  \\
    \midrule
    \multicolumn{7}{c}{\textit{Multiple Proposals Only}}\\
    \midrule
    GroundingDINO-B & 48.5 & 48.9 & 47.1 & 45.1 & 49.1 & 51.5  \\
    + \llavasmall{} & 62.0 & 50.0 & 60.5 & \textbf{47.9} & 57.2 & 54.6  \\
    + \llavasmall{} + \METHOD{} (Ours) & \textbf{66.9} & \textbf{51.4} &  \textbf{64.8} & 47.0 & \textbf{61.1} & \textbf{60.4}  \\
    \bottomrule
    \end{tabular}
    }
\end{table}

\textbf{\METHOD{} improves region-text alignment in VLMs.}
In \cref{tab:visual_grounding}, \METHOD{} surpasses GroundingDINO in terms of top prediction accuracy and demonstrates superior performance over using \llava{}'s probabilities in all scenarios except for the RefCOCO+ testB. 
We improve by $2.73\%$ on average compared to the top prediction by GroundingDINO, and by $0.8\%$ on average over re-ranking with \llavasmall{}.
It is important to note that in scenarios where only a single bounding box is available, re-ranking is infeasible, and we select the single box by default. 
Thus, we additionally show the results on the subset of data where there are multiple proposals, which on average accounts for $28.6\%$ of the data. 
As illustrated in the bottom section of \cref{tab:visual_grounding}, \METHOD{} reveals a much larger improvement, \eg{}, $3.2\%$, $1.7\%$, and $3.9\%$ on average for the RefCOCO, RefCOCO+, and RefCOCOg test splits, and a $5.8\%$ improvement for the Flickr30K Entities test set. This indicates the value of our approach in linking phrases to the most pertinent image regions.

\subsection{Analysis and Ablation Studies}
\label{sec:ablation}

In the following, we analyze design choices for \METHOD{}, including the 
comparison of different region guidance strategies (\cref{sec:ablation_whatsup}), analyzing probability shift for grounded text to understand why \METHOD{} works (\cref{sec:why_crg_works_shift_prob}), and the impact of region guidance strength $\alpha$ (\cref{sec:ablation_alpha}). 

\begin{figure}[tb]
    \centering
    \textit{Blacking out different Regions}
    
    \begin{subfigure}{.245\linewidth}
      \includegraphics[width=\linewidth]{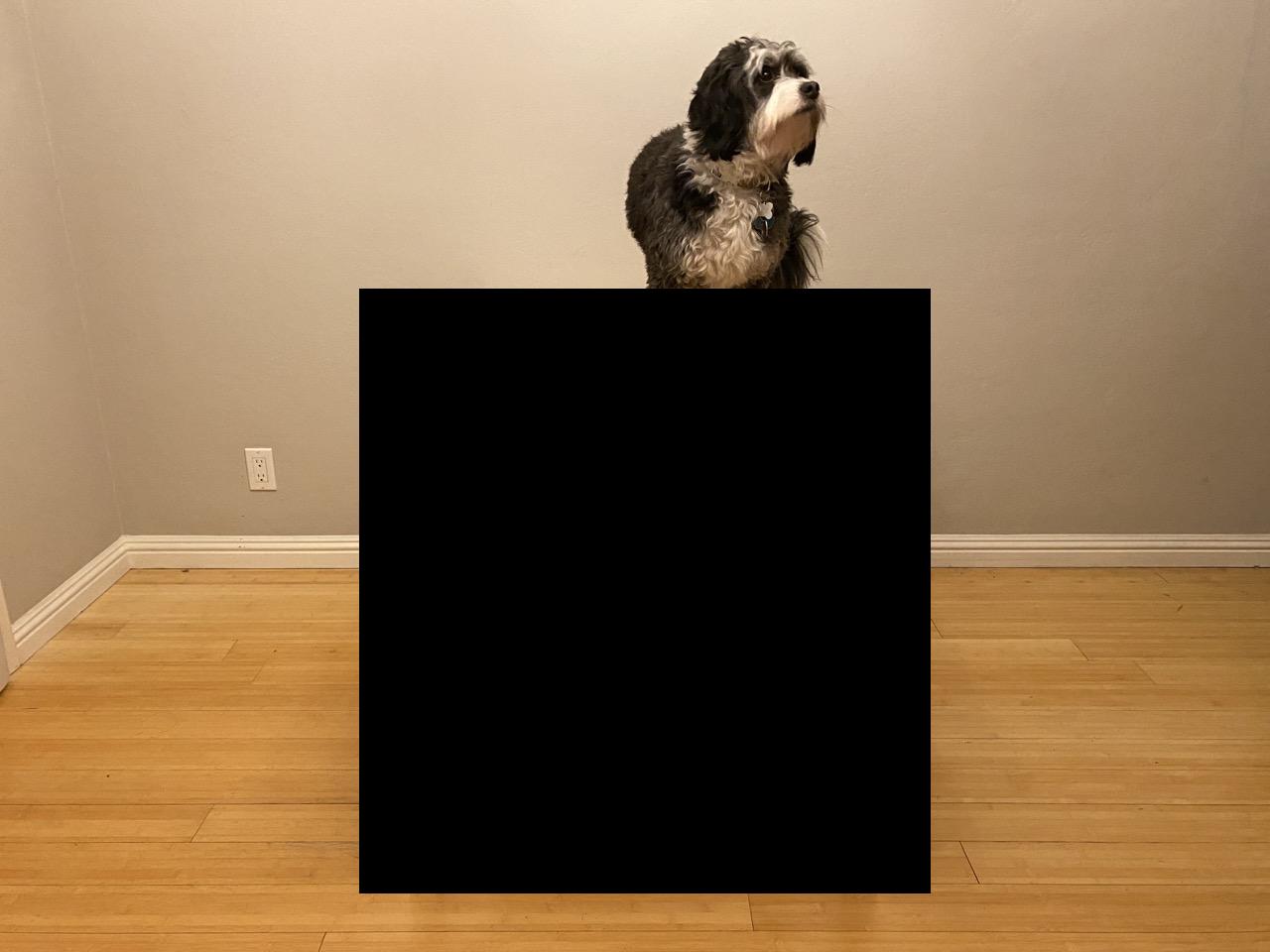}
      \caption{Blackout Single Object}
    \end{subfigure}%
    \hfill
    \begin{subfigure}{.245\linewidth}
      \includegraphics[width=\linewidth]{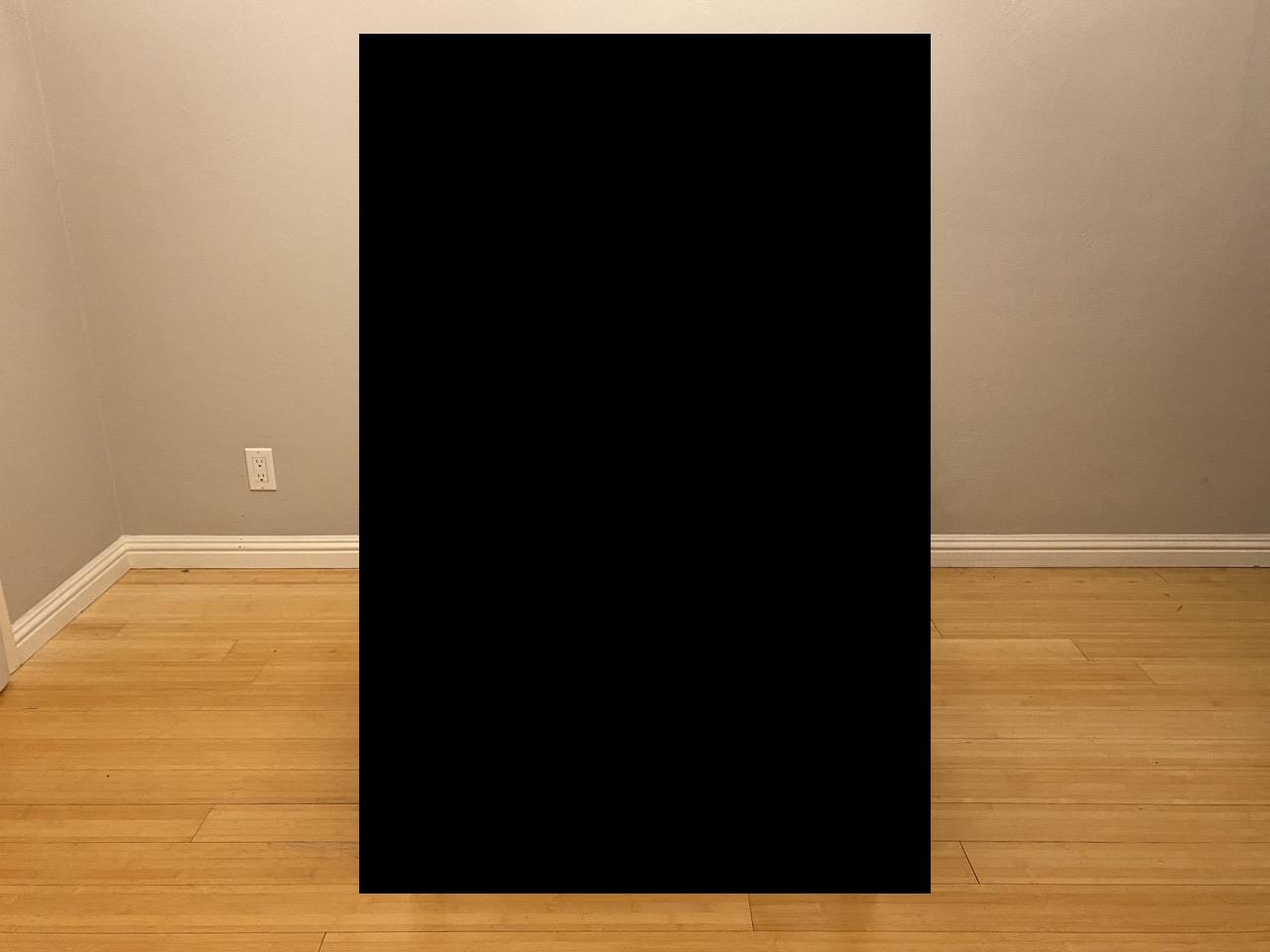}
      \caption{Blackout Combined}
    \end{subfigure}%
    \hfill
    \begin{subfigure}{.245\linewidth}
      \includegraphics[width=\linewidth]{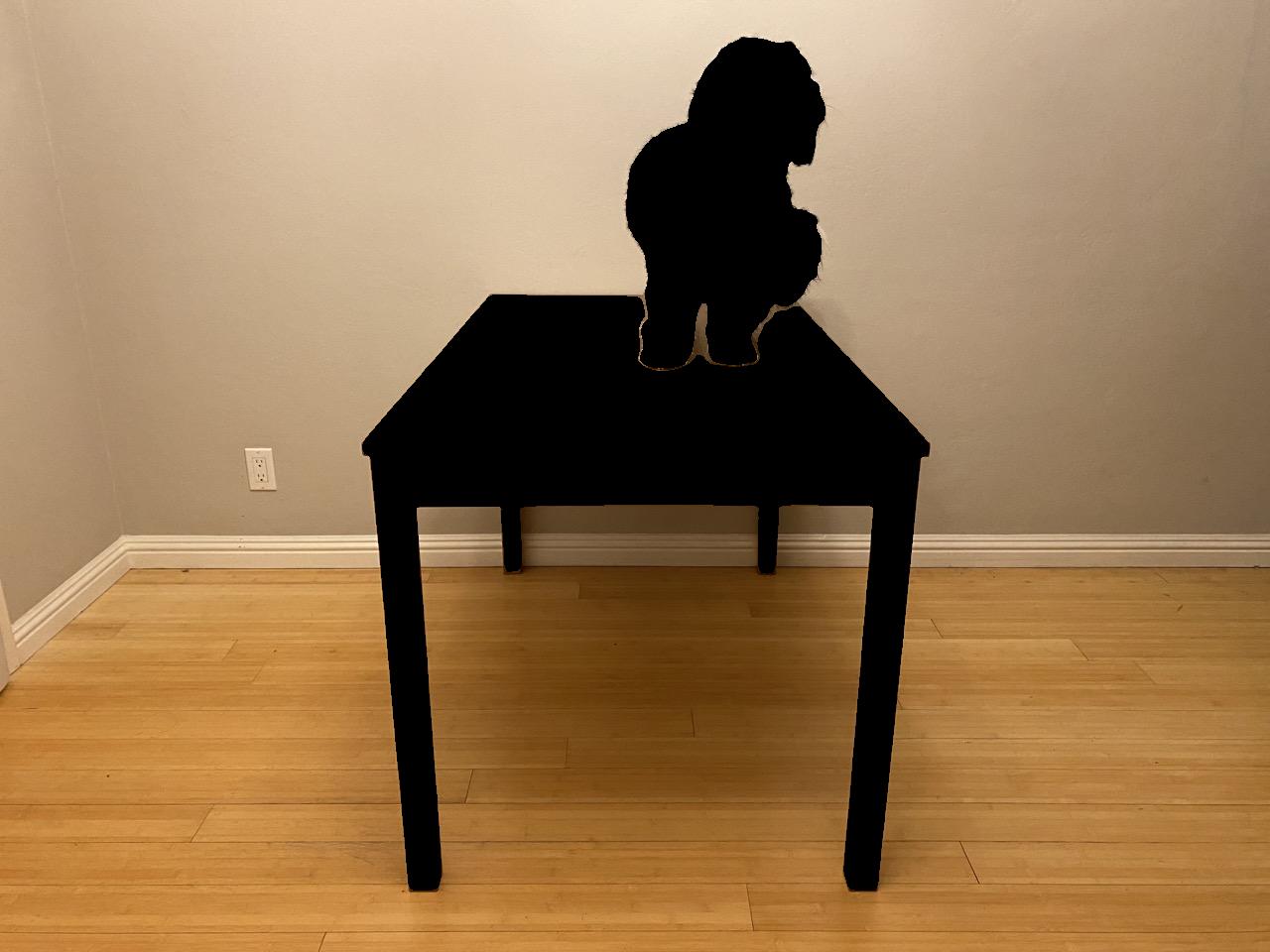}
      \caption{Blackout with Mask}
    \end{subfigure}%
    \hfill
    \begin{subfigure}{.245\linewidth}
      \includegraphics[width=\linewidth]{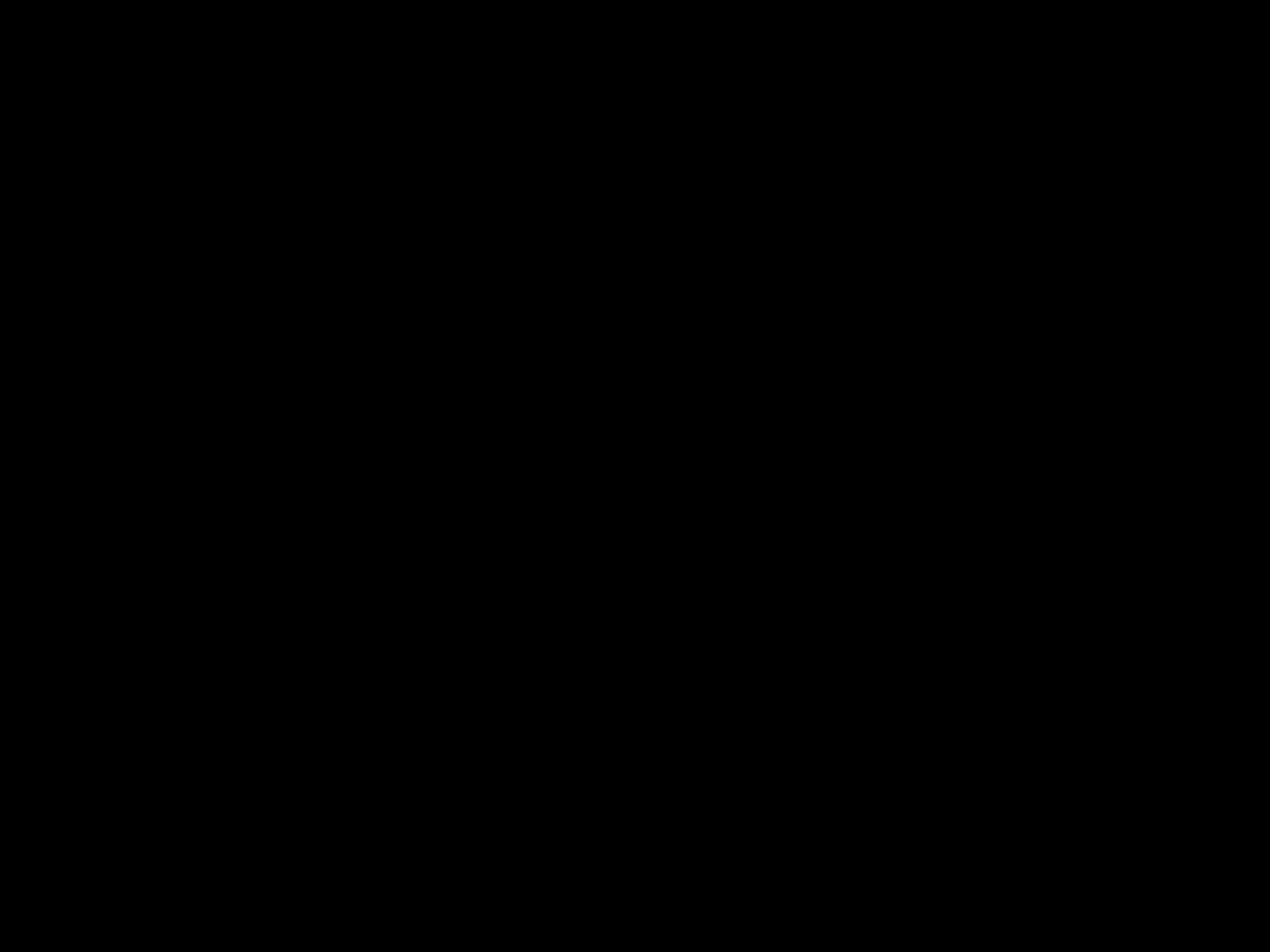}
      \caption{Blackout All}
    \end{subfigure}
    
    \textit{\METHOD{} vs. Overlay}
    
    \begin{subfigure}{.245\linewidth}
      \includegraphics[width=\linewidth]{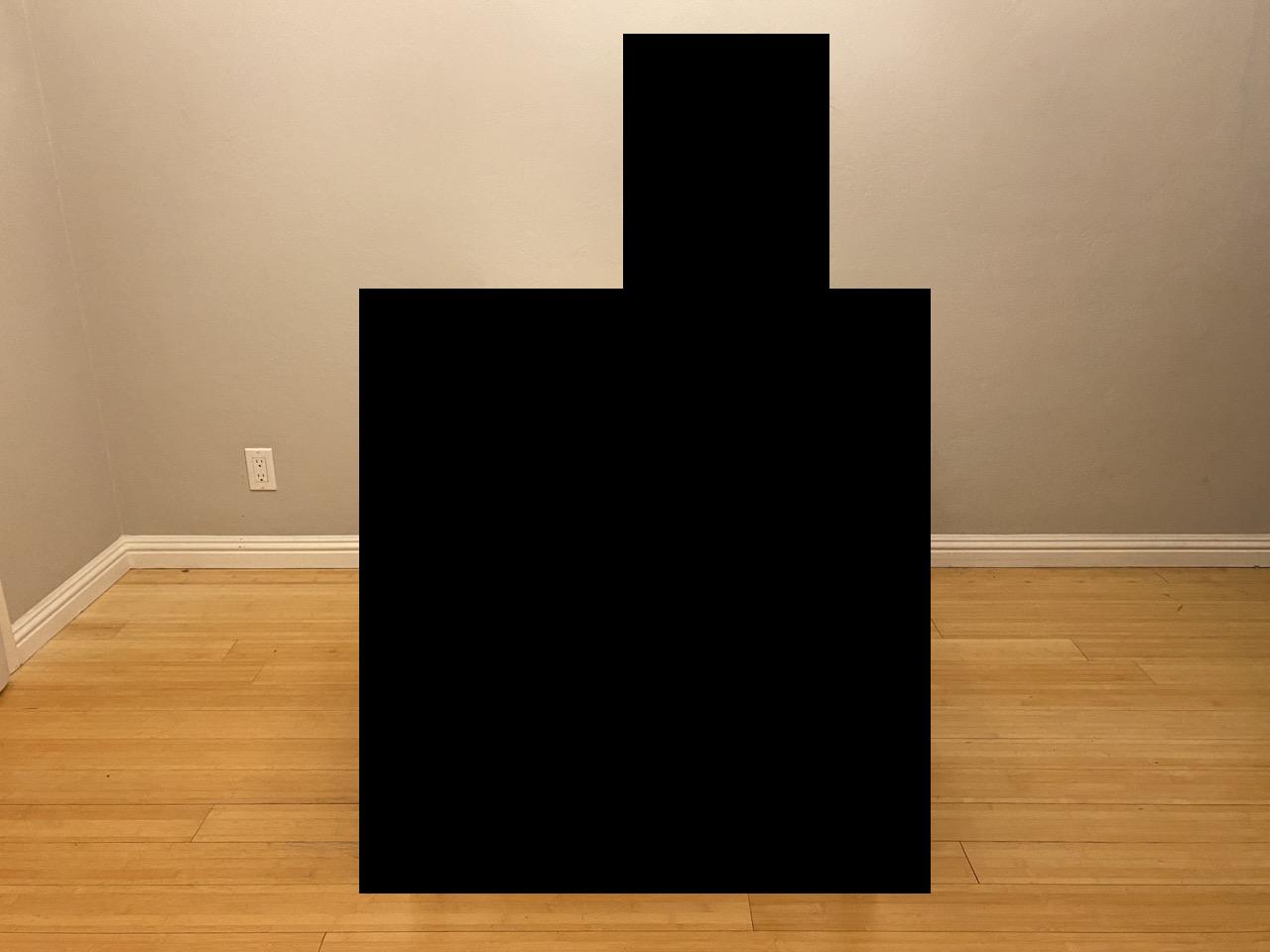}
      \caption{Blackout \METHOD{} }
    \end{subfigure}%
    \hfill
    \begin{subfigure}{.245\linewidth}
      \includegraphics[width=\linewidth]{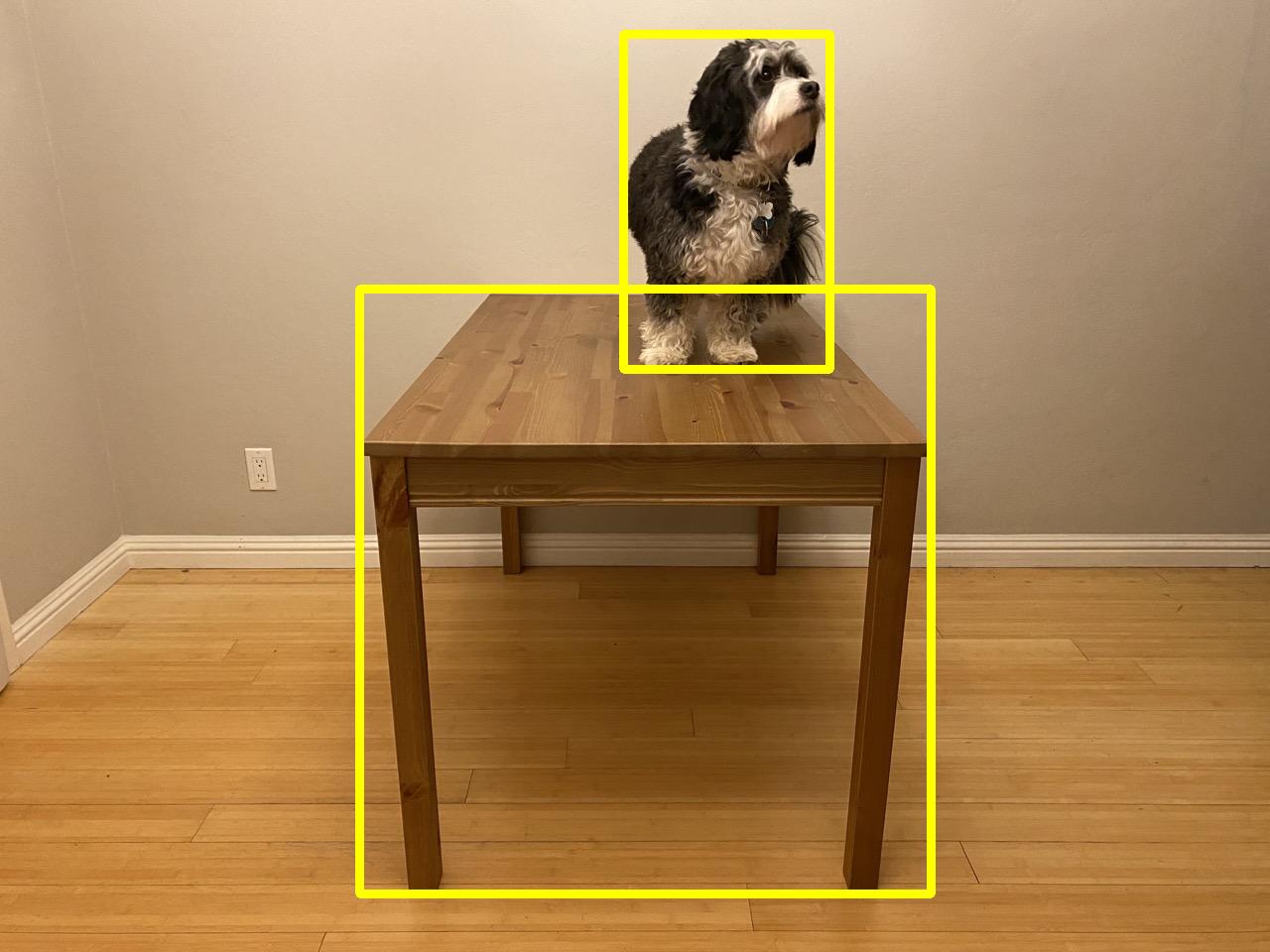}
      \caption{Overlay with Boxes}
    \end{subfigure}%
    \hfill
    \begin{subfigure}{.245\linewidth}
      \includegraphics[width=\linewidth]{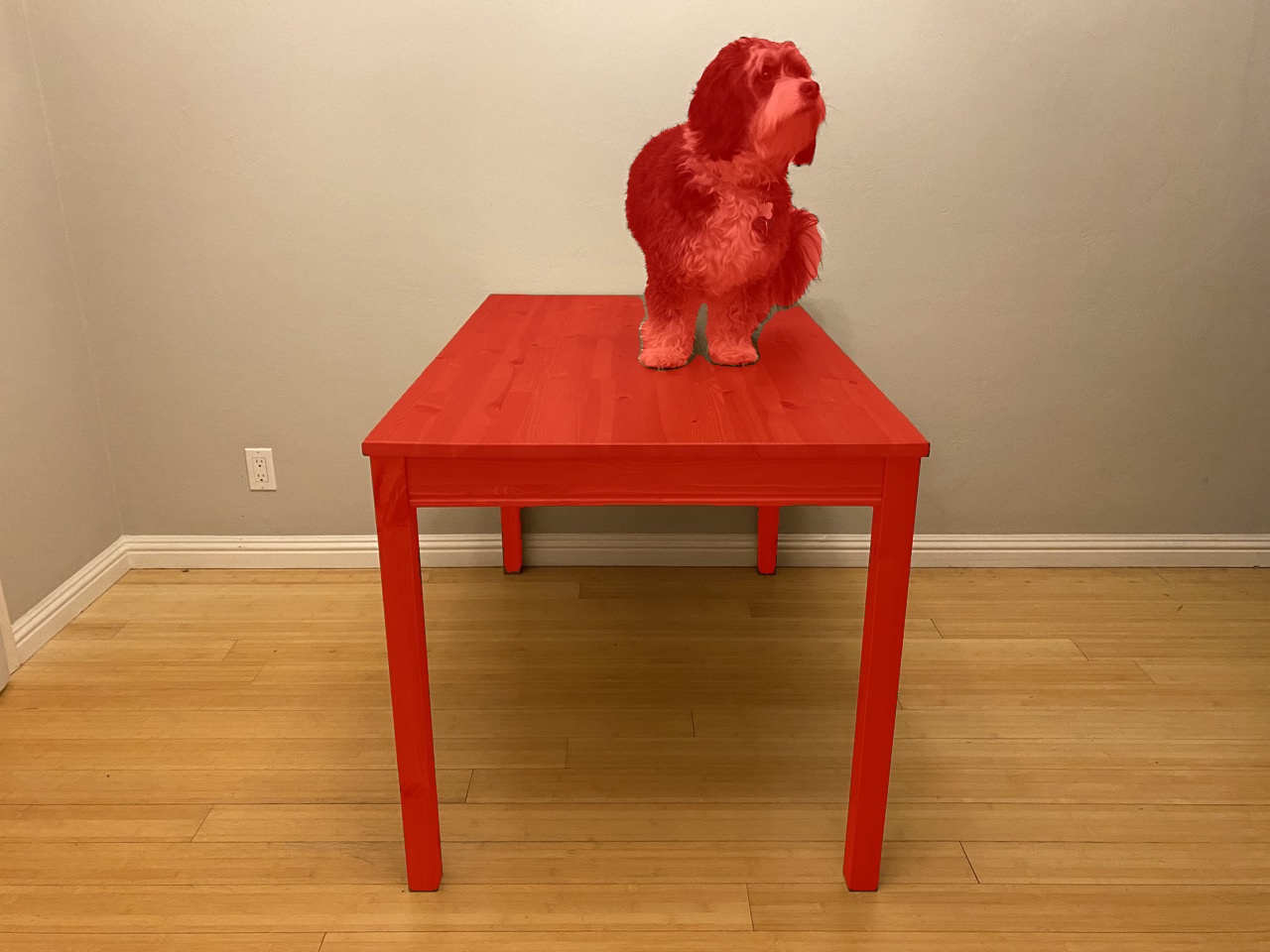}
      \caption{Overlay with Masks}
    \end{subfigure}%
    \hfill
    \begin{subfigure}{.245\linewidth}
      \includegraphics[width=\linewidth]{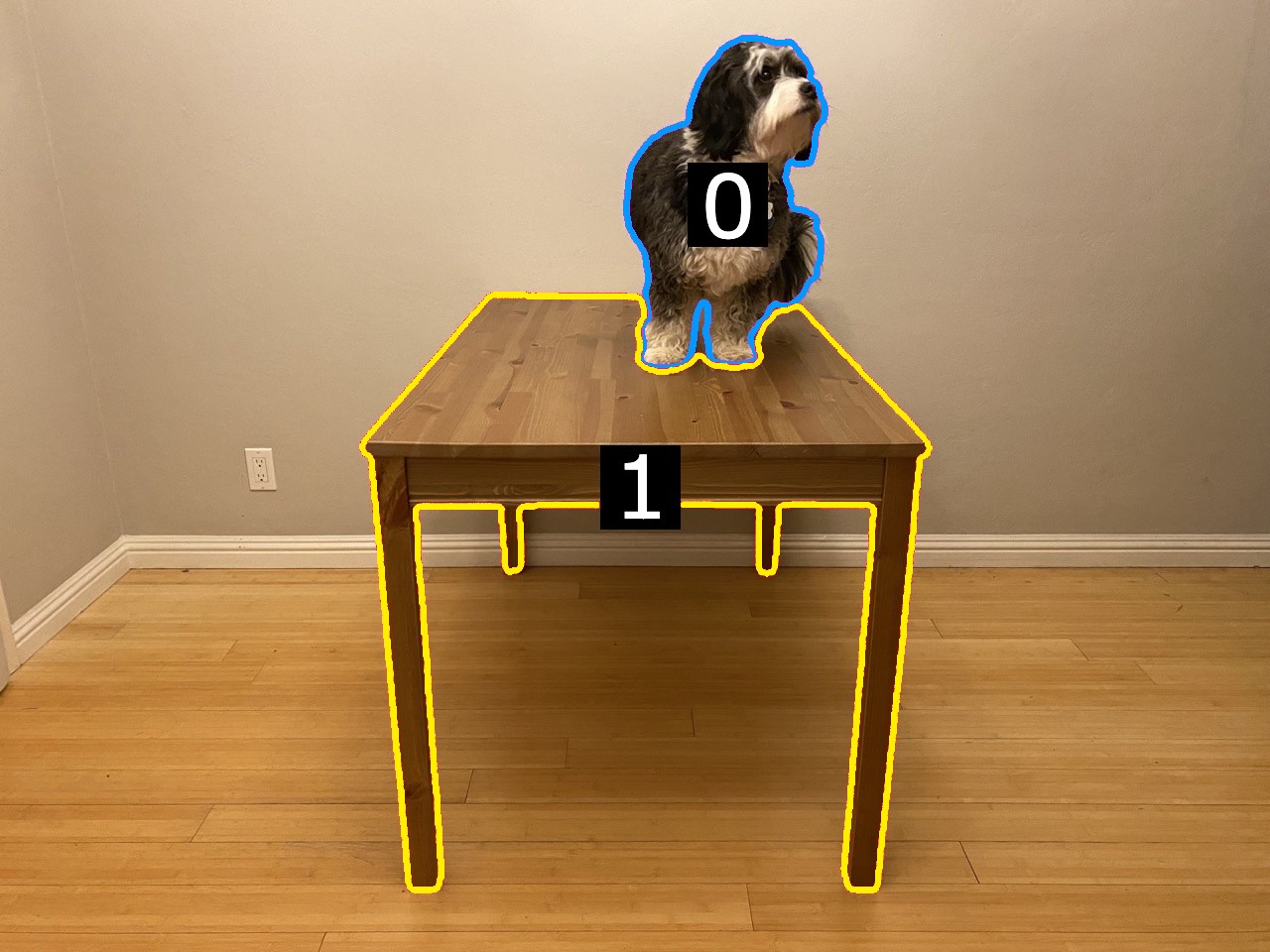}
      \caption{Overlay with SoM}
    \end{subfigure}
    \caption{Different masking and overlaying strategies on the \whatsup{} dataset containing two objects.
    \textbf{Top}: Blacking out with different regions.
    \textbf{Bottom}: \METHOD{}'s blackout strategy (e) contrasted with different methods for overlaying visual markers (f-h).
    }
    \label{fig:black_out_type}
\end{figure}

\begin{table}[tb]
    \caption{\whatsup{} results with different region guidance strategies on \llavalarge{}.}
    \label{tab:whatsup_ablations}
    \small
    \centering
    \begin{tabular}{l c c c}
    \toprule
    Method & Indiv. & Pairs & Set of 4 \\
    \midrule
    Original Image &  86.8 &  76.8 &  54.0 \\
    \midrule
    \multicolumn{4}{c}{\textit{Contrasted with Original Image}}\\
    \midrule
    (a) Blackout Single Object & 86.7 & 78.1 & 59.0 \\
    (b) Blackout Combined &  85.8 & 78.6 & 60.1 \\
    (c) Blackout w. Mask &  85.4  & 75.4 & 55.7 \\
    (d) Blackout All &  82.6 &  75.2 & 53.8\\
    (e) Blackout \METHOD{} & \textbf{87.6} & \textbf{80.3}  & \textbf{64.0} \\
    \midrule
    \multicolumn{4}{c}{\textit{Overlay on Original Image}}\\
    \midrule
    (f) Overlay with Boxes & 82.2 & 69.4 & 39.7  \\
    (g) Overlay with Masks & 84.2  & 82.7 & 49.6  \\
    (h) Overlay with SoM & 77.9 & 63.9 & 33.0 \\

    \bottomrule
    \end{tabular}
\end{table}

\subsubsection{Different Region Guidance Strategies}
\label{sec:ablation_whatsup}

We investigate the impact of different region guidance strategies, including
contrasting original images with another image (\eg{}, images where different regions are blacked out) 
and overlaying visual markers (\eg{}, bounding box and segmentation mask)
on the \whatsup{} benchmark, given that each scene in the dataset contains exactly two distinct objects for reliable analysis. 
Here, we use \llavalarge{}.
As depicted in the top section of \cref{fig:black_out_type}, apart from our method of applying bounding boxes for each object separately shown in (e), we apply four distinct masking approaches. First, considering the different combinations of the objects, we black out one of the objects (we take the average of the results of masking each object) in (a), and apply a combined mask over both objects in (b).
We also consider blacking out with a segmentation mask using Grounded-SAM \cite{ren2024grounded} in \cref{fig:black_out_type} (c), motivated by the success of overlaying such masks as visual prompts \cite{Yang2023SoM}, and blacking the entire image in (d) as ablations, which has been previously applied to CFG \cite{Kornblith2023CFGCaption}.
We are primarily interested in finding the best strategy to mask especially in the presence of multiple objects, as well as the effect of the granularity of the mask.

\textbf{Blacking out only the relevant regions is important.}
The findings are detailed in \cref{tab:whatsup_ablations}. 
Our method of masking each object separately achieves superior performance compared to the other masking strategies. In particular, our method (e) performs better than blacking out a combined mask (b) and blacking out the entire image (d), indicating the importance of precisely targeting necessary regions for removal to prevent the accidental exclusion of additional information. 
When blacking out with a segmentation mask in (c), we observe that while this approach yields competitive results among other masking strategies, it is still worse than the our main method of (Blackout Separate). This indicates that the model may be using object's shape, which segmentation masks preserve. 

\textbf{Simply overlaying visual markers without CRG is ineffective for pre-trained VLMs.}
Finally, we explore different visual markers for direct visual prompting, including bounding boxes (f) and segmentation masks (g), which has been shown to be effective for trained models \cite{sun2023alphaclip, Cai2023VIP-LLAVA}, in \cref{fig:black_out_type}.
Lastly, we also experiment with SoM \cite{Yang2023SoM} in (h). 
The results in \cref{tab:whatsup_ablations} show that the \llavalarge{} model does not follow such visual prompts, as the three overlaying methods show worse performance than using the original image. The decrease in performance with SoM also echoes the result in \cref{tab:vip_bench}.
As mentioned in \cref{sec:method}, overlaying does not guarantee that the model will focus on the region of interest -- or contrast with it -- as it can still attend to any spurious information outside that region, as in the original image. 
Contrasting with a blacked-out image reduces this kind of spurious information by factoring it out when the logits are subtracted. Thus, we show the usefulness of our chosen blackout strategy in aiding the model to follow visual prompts.

\subsubsection{Quantifying the Intuition Behind \METHOD{}: Probability Contrasts for Grounded Text} 
\label{sec:why_crg_works_shift_prob}

\begin{figure}[tb]
    \centering
    \begin{subfigure}{.59\linewidth}
      \includegraphics[height=0.18\textheight]{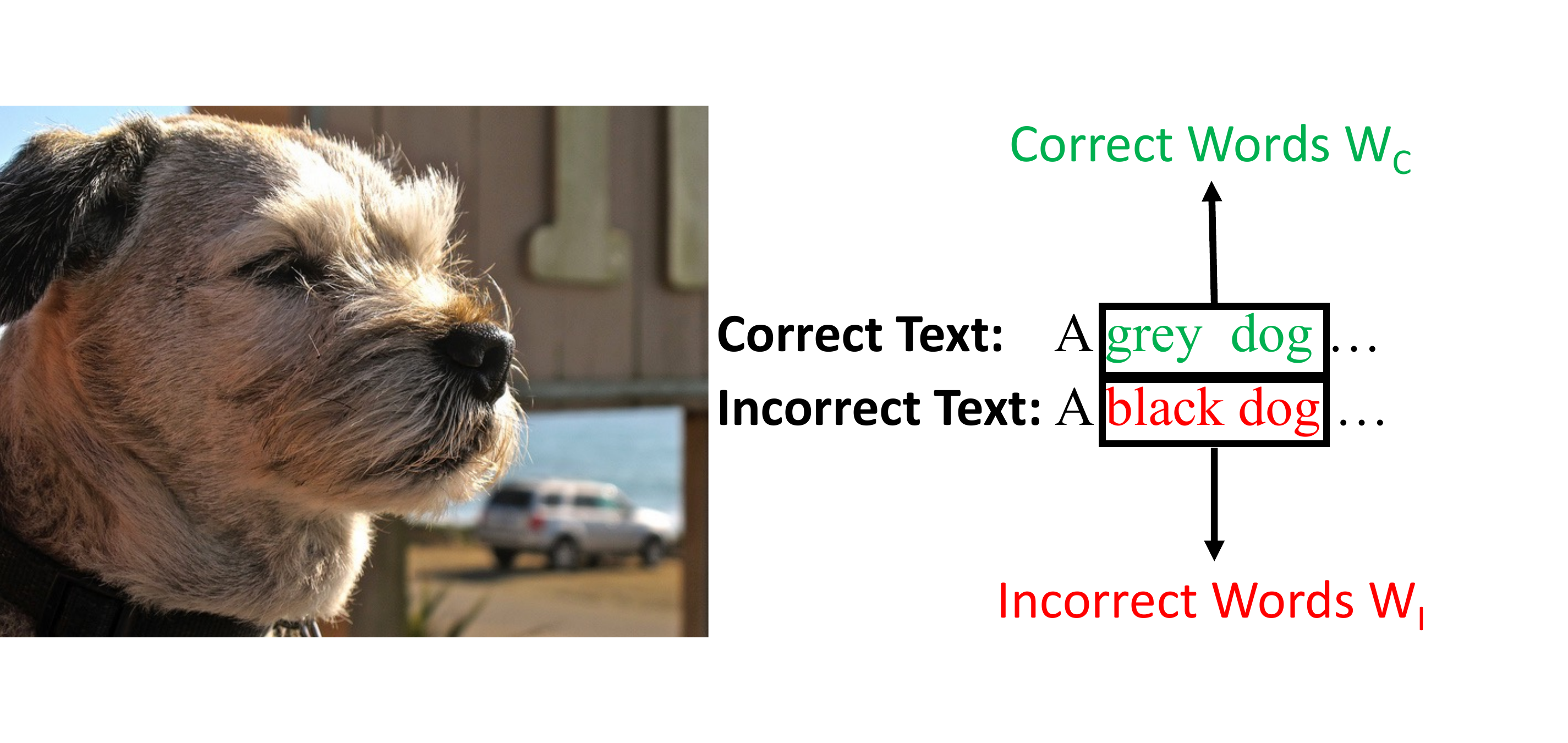}
      \caption{}
    \end{subfigure}%
    \hfill
    \begin{subfigure}{.39\linewidth}
      \includegraphics[height=0.18\textheight]{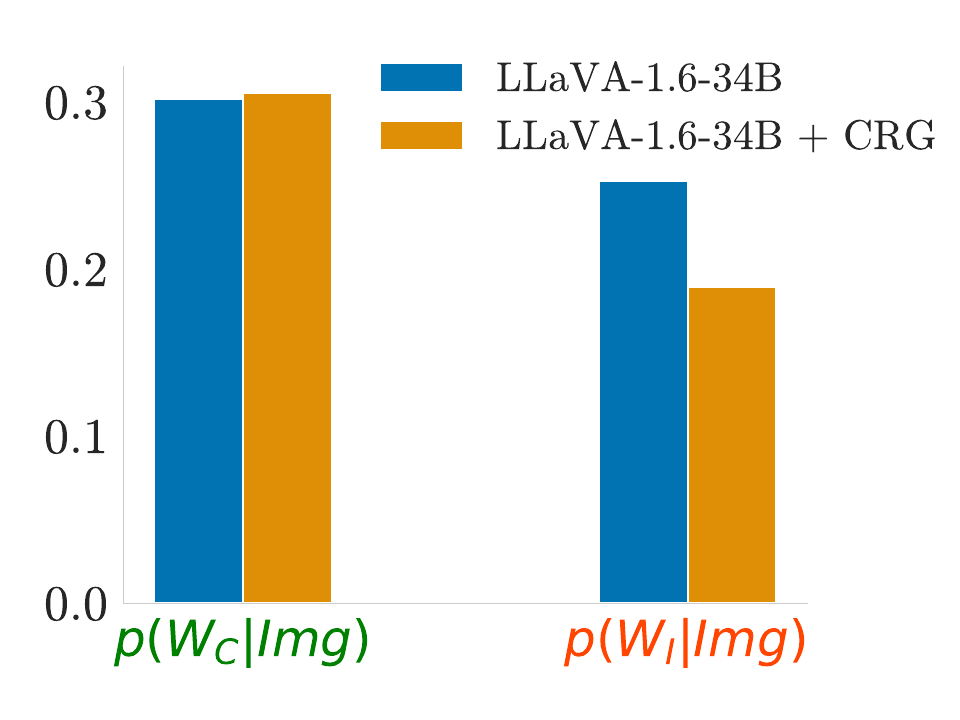}
      \caption{}
    \end{subfigure}
    \caption{(a) shows an example of correct and incorrect texts from \sugarcrepe{} \swapatt{}. The correct text contains correct words $W_{C}$ (\ie{}, grey dog) that are reflected in the image, whereas the incorrect text has the attribute swapped to form the incorrect words $W_{i}$ (\ie{}, black dog). We compare the average probability assigned to all correct words $W_{C}$ and all incorrect words $W_{I}$  by \llavalarge{} and \llavalarge{} + \METHOD{} in (b). 
    }
    \label{fig:sugarcrepe_ablations}
\end{figure}

To better understand \emph{why} \METHOD{} improves visual grounding, we analyze \METHOD{}'s behavior on \sugarcrepe{}, which measures compositional generalization.
Here, we examine how the probability distribution on key words tied to the proposed regions changes when applying \METHOD{}.
In \sugarcrepe{}, each image has one correct caption and an incorrect distractor caption. 
In \swapatt{}, the distractors are formed by swapping the visual attributes of objects in the scene; for example in \cref{fig:sugarcrepe_ablations} (a), ``\textit{grey dog}'' is changed to ``\textit{black dog}''.
We treat the attribute phrases from the correct captions as correct words $W_{C}$ (\eg{}, ``grey dog'') and the phrase from incorrect captions as incorrect words $W_{I}$ (\eg{}, ``black dog'').
If \METHOD{} operates as expected, \METHOD{} should increase the model's probability on \emph{correct words} $W_{C}$, as \METHOD{} emphasizes the correct object in the image while lowering the probability of incorrect words $W_{I}$, which cannot be inferred from the regions.

\paragraph{\METHOD{} amplifies the correct text probability and reduces incorrect text probability.}
We use \llavalarge{} and calculate the average probabilities for the $W_{C}$ and $W_{I}$.
We compare the probabilities obtained from \llavalarge{} alone to those obtained by applying \METHOD{} in \cref{fig:sugarcrepe_ablations} (b).
We see that the probability of the correct phrases increases slightly, while the probability of the incorrect phrase decreases with \METHOD{}. 
This indicates that the model is following the visual prompts and paying attention to the correct image regions to better differentiate the positive text from the negative, and is able to ground these regions in the image to the relevant part of the text. 
Thus, \METHOD{} improves performance in an \emph{interpretable} way, by improving the matching between the image and the relevant tokens in the text. 

\begin{figure}[tb]
    \centering
    \begin{subfigure}{.32\linewidth}
      \includegraphics[width=\linewidth]{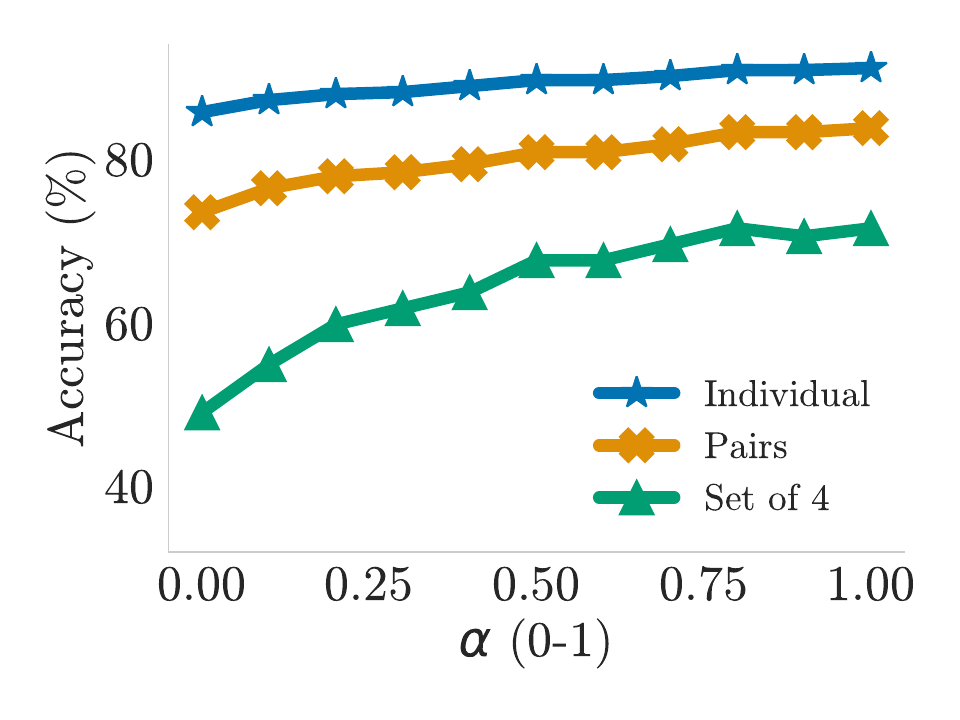}
    \end{subfigure}%
    \hfill
    \begin{subfigure}{.32\linewidth}
      \includegraphics[width=\linewidth]{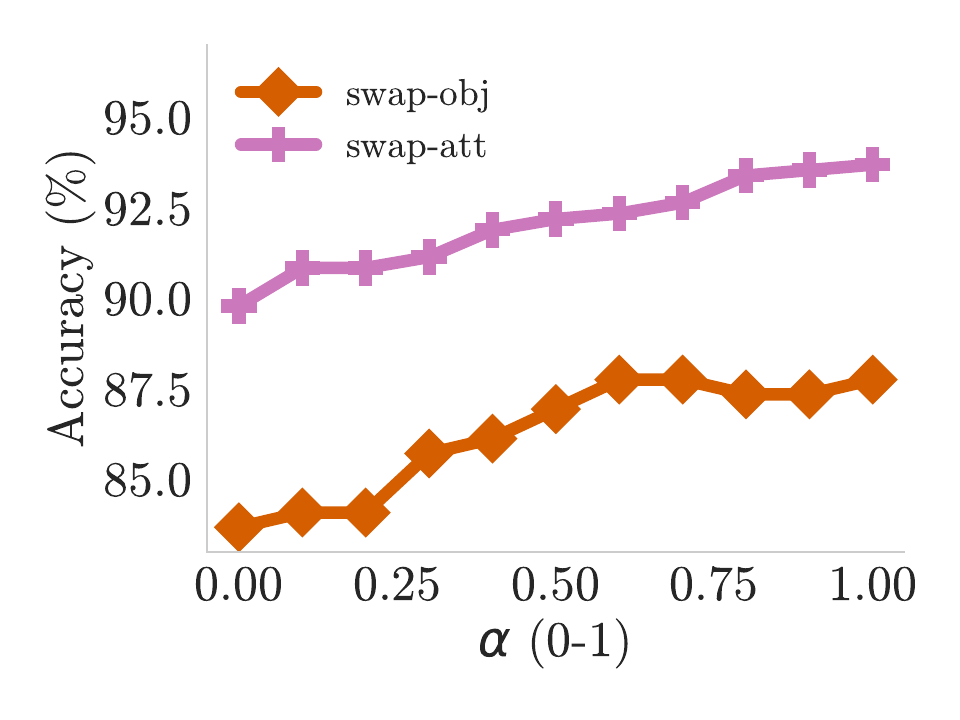}
    \end{subfigure}%
    \hfill
    \begin{subfigure}{.32\linewidth}
      \includegraphics[width=\linewidth]{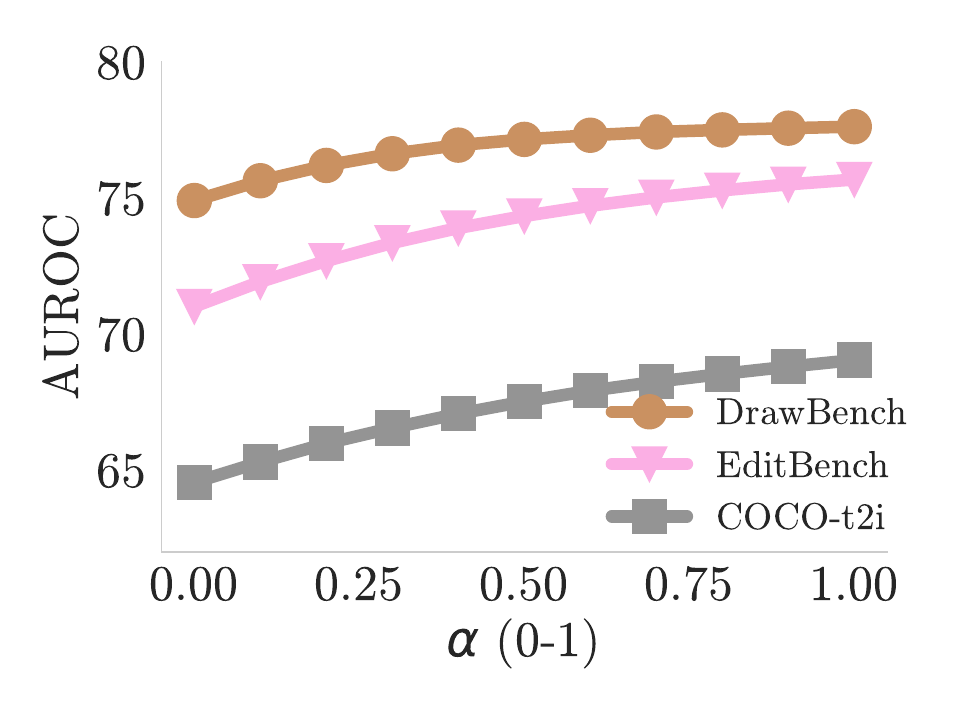}
    \end{subfigure}%
    \hfill
    \begin{subfigure}{.32\linewidth}
      \includegraphics[width=\linewidth]{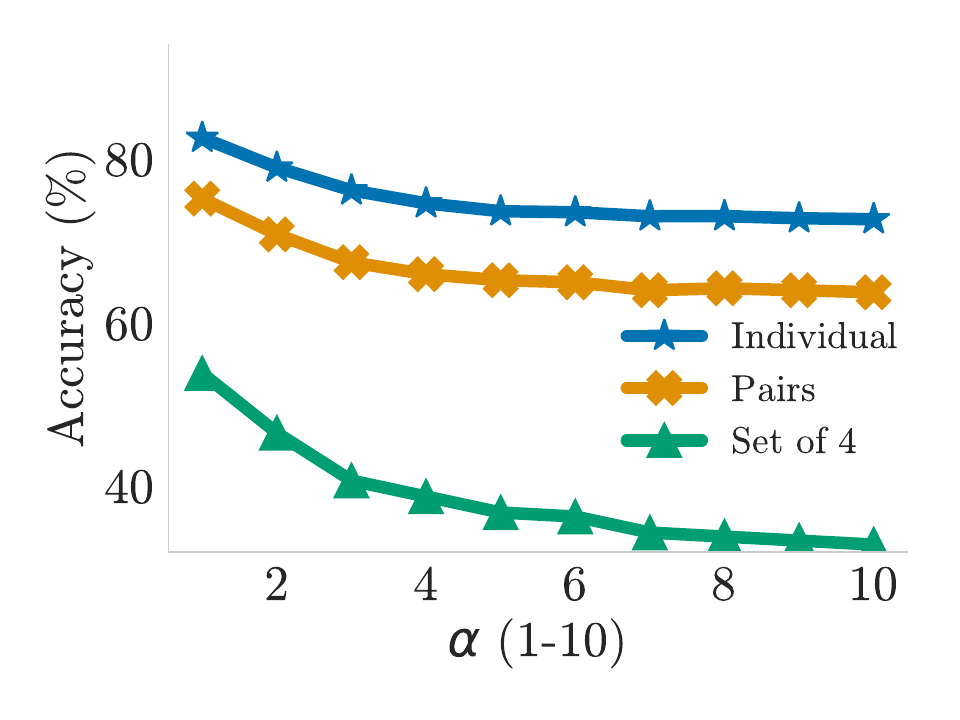}
      \caption{\whatsup{}}
    \end{subfigure}%
    \hfill
    \begin{subfigure}{.32\linewidth}
      \includegraphics[width=\linewidth]{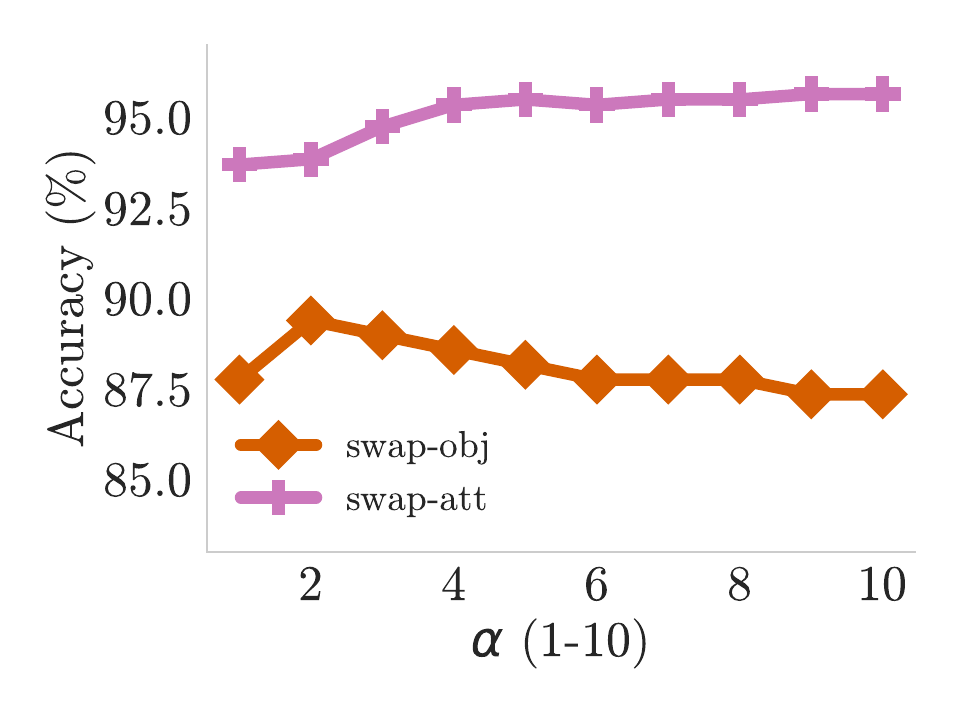}
      \caption{\sugarcrepe{}}
    \end{subfigure}%
    \hfill
    \begin{subfigure}{.32\linewidth}
      \includegraphics[width=\linewidth]{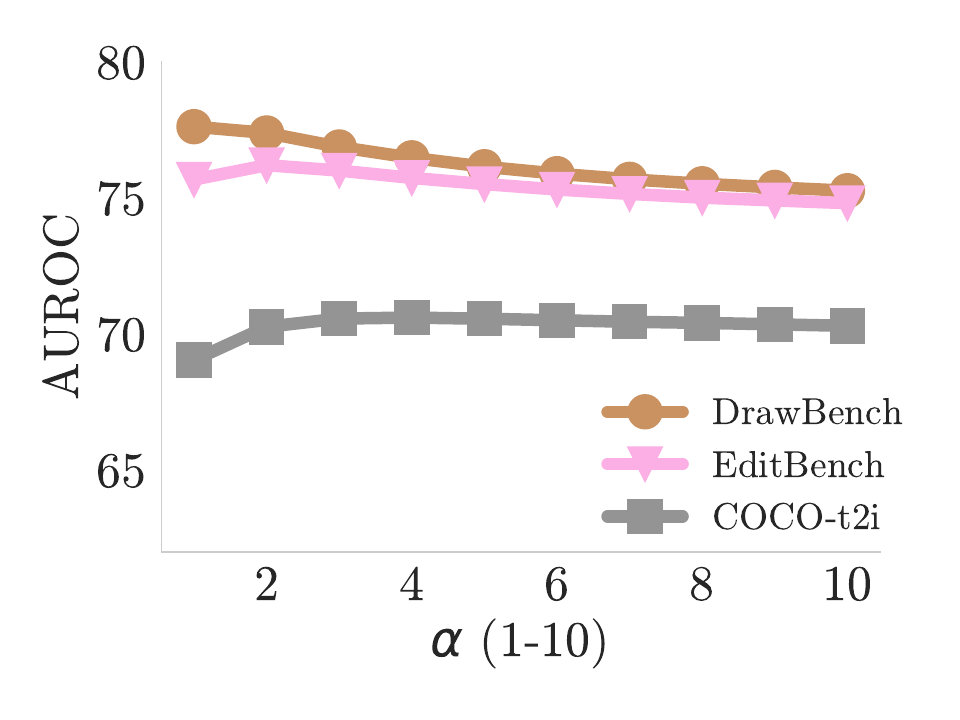}
      \caption{\seetrue{}}
    \end{subfigure}%
    \hfill
    \caption{Ablations of $\alpha$ on \whatsup{} (a), \sugarcrepe{} (b), and \seetrue{} (c). We evaluate $\alpha$ between 0 and 1 on the top graphs, and between 1 and 10 on the bottom graphs.
    }
    \label{fig:alpha_ablations}
\end{figure}

\subsubsection{Impact of Region Guidance Strength \texorpdfstring{$\alpha$}{a}}\label{sec:ablation_alpha}

We analyze the impact of the region guidance strength $\alpha$ on various tasks.
As explained in \cref{sec:method},
$\alpha$ in the~\cref{eq:cfg}: $\texttt{softmax}[ (1 + \alpha) \cdot \text{logit}_{\theta}(y_{t} | I,X,y_{<t}) - \alpha \cdot \text{logit}_{\theta}(y_t|\texttt{mask}(I,b),X,y_{<t})]$ controls how strongly \METHOD{} guides VLMs to focus on the region $b$.
We illustrate the effect of adjusting $\alpha$ from 0 (regular decoding from the original image) to 1 with a step of 0.1. \cref{fig:alpha_ablations} shows the accuracy (and AUROC for \seetrue{}) across different datasets with various $\alpha$. 
We observe a clear trend that increasing $\alpha$ from 0 to 1 improves performance, indicating that focusing on the provided regions is more beneficial to the tasks.
Based on this finding, we further experiment with increasing the $a$ value up to 10, which is shown in the bottom section of \cref{fig:alpha_ablations}. While for some tasks, such as \sugarcrepe{}, setting a more aggressive weight improves performance there is no clear trend where a single value can achieve the best performance. Thus, we advocate for $\alpha=1$ as a default value that can be optionally tuned if a validation set is present.

\section{Conclusion}
We present \METHOD{}, an accessible and easy-to-use training-free approach for improving visual prompt following capability for VLMs without the need for training.
\METHOD{} provides significant improvement in visual prompt following and is effective across a broad spectrum of vision-language tasks that lack ground truth for region annotations: \METHOD{} improves spatial reasoning, compositional generalization, and image-text alignment for generated images by employing a re-ranking strategy for regions identified by an object detection model. 
We further explore different region guidance strategies, aiming to set a foundation for future advancements in visual prompting techniques. One direction for future work includes the integration of visual and textual contexts: our research focuses on guiding the model with visual inputs, and a concurrent work \cite{zhao2024mitigating} proposes guidance through the added textual context in the caption. 
We believe that these directions complement each other and suggest a combined approach for an enhanced multimodal prompt following strategy.

\section*{Limitations}
\METHOD{} shows strong performance across a variety of VL tasks; however, like all CFG methods, it comes with an additional computational cost due to the necessity of running the model twice (on the original image and the blacked-out image). 
This cost is offset by the fact that \METHOD{} is broadly applicable across various models and datasets, and does not require fine-tuning -- although it can also be complementary to fine-tuned models, as shown in \cref{sec:vip-bench}.
If visual markers are absent, \METHOD{} also relies on an object detection model; however, such models are available across many domains.
Future work could incorporate better visual encoders that can yield direct identification of relevant regions without additional object detectors.

\section*{Acknowledgement}
We thank Peter Hase for the thoughtful discussion. This work was supported by DARPA ECOLE Program No. HR00112390060, NSF-AI Engage Institute DRL-2112635, DARPA Machine Commonsense (MCS) Grant N66001-19-2-4031, ARO Award W911NF2110220, ONR Grant N00014-23-1-2356, and a Bloomberg Data Science Ph.D. Fellowship. 
The views contained in this article are those of the authors and not of the funding agency.

%%%%%%%% NeurIPS FORMAT BIB
\bibliographystyle{abbrvnat}
\bibliography{main}

\begin{thebibliography}{63}
\providecommand{\natexlab}[1]{#1}
\providecommand{\url}[1]{\texttt{#1}}
\expandafter\ifx\csname urlstyle\endcsname\relax
  \providecommand{\doi}[1]{doi: #1}\else
  \providecommand{\doi}{doi: \begingroup \urlstyle{rm}\Url}\fi

\bibitem[Antol et~al.(2015)Antol, Agrawal, Lu, Mitchell, Batra, Zitnick, and Parikh]{antol2015vqa}
S.~Antol, A.~Agrawal, J.~Lu, M.~Mitchell, D.~Batra, C.~L. Zitnick, and D.~Parikh.
\newblock Vqa: Visual question answering.
\newblock In \emph{Proceedings of the IEEE international conference on computer vision}, pages 2425--2433, 2015.

\bibitem[Bahng et~al.(2022)Bahng, Jahanian, Sankaranarayanan, and Isola]{bahng2022exploring}
H.~Bahng, A.~Jahanian, S.~Sankaranarayanan, and P.~Isola.
\newblock Exploring visual prompts for adapting large-scale models, 2022.

\bibitem[Bai et~al.(2023{\natexlab{a}})Bai, Bai, Yang, Wang, Tan, Wang, Lin, Zhou, and Zhou]{Qwen-VL}
J.~Bai, S.~Bai, S.~Yang, S.~Wang, S.~Tan, P.~Wang, J.~Lin, C.~Zhou, and J.~Zhou.
\newblock Qwen-vl: A versatile vision-language model for understanding, localization, text reading, and beyond.
\newblock \emph{arXiv preprint arXiv:2308.12966}, 2023{\natexlab{a}}.

\bibitem[Bai et~al.(2023{\natexlab{b}})Bai, Geng, Mangalam, Bar, Yuille, Darrell, Malik, and Efros]{bai2023sequentialLVM}
Y.~Bai, X.~Geng, K.~Mangalam, A.~Bar, A.~Yuille, T.~Darrell, J.~Malik, and A.~A. Efros.
\newblock Sequential modeling enables scalable learning for large vision models, 2023{\natexlab{b}}.

\bibitem[Bar et~al.(2022)Bar, Gandelsman, Darrell, Globerson, and Efros]{Bar2022}
A.~Bar, Y.~Gandelsman, T.~Darrell, A.~Globerson, and A.~A. Efros.
\newblock {Visual Prompting via Image Inpainting}.
\newblock In \emph{NeurIPS}, 2022.
\newblock ISBN 9781713871088.

\bibitem[Cai et~al.(2023)Cai, Liu, Mustikovela, Meyer, Chai, Park, and Lee]{Cai2023VIP-LLAVA}
M.~Cai, H.~Liu, S.~K. Mustikovela, G.~P. Meyer, Y.~Chai, D.~Park, and Y.~J. Lee.
\newblock {Making Large Multimodal Models Understand Arbitrary Visual Prompts}, 2023.
\newblock URL \url{http://arxiv.org/abs/2312.00784}.

\bibitem[Chen et~al.(2023{\natexlab{a}})Chen, Zhang, Zeng, Zhang, Zhu, and Zhao]{chen2023shikra}
K.~Chen, Z.~Zhang, W.~Zeng, R.~Zhang, F.~Zhu, and R.~Zhao.
\newblock Shikra: Unleashing multimodal llm's referential dialogue magic.
\newblock \emph{arXiv preprint arXiv:2306.15195}, 2023{\natexlab{a}}.

\bibitem[Chen et~al.(2022)Chen, Zhao, Zhang, Duan, Qi, and Zhao]{focalclick}
X.~Chen, Z.~Zhao, Y.~Zhang, M.~Duan, D.~Qi, and H.~Zhao.
\newblock Focalclick: Towards practical interactive image segmentation.
\newblock In \emph{Proceedings of the IEEE/CVF Conference on Computer Vision and Pattern Recognition (CVPR)}, pages 1300--1309, June 2022.

\bibitem[Chen et~al.(2023{\natexlab{b}})Chen, Djolonga, Padlewski, Mustafa, Changpinyo, Wu, Ruiz, Goodman, Wang, Tay, et~al.]{chen2023pali}
X.~Chen, J.~Djolonga, P.~Padlewski, B.~Mustafa, S.~Changpinyo, J.~Wu, C.~R. Ruiz, S.~Goodman, X.~Wang, Y.~Tay, et~al.
\newblock Pali-x: On scaling up a multilingual vision and language model.
\newblock \emph{arXiv preprint arXiv:2305.18565}, 2023{\natexlab{b}}.

\bibitem[Cho et~al.(2023)Cho, Kim, Ryu, and Kweon]{cho2023generative}
J.~W. Cho, D.-J. Kim, H.~Ryu, and I.~S. Kweon.
\newblock Generative bias for robust visual question answering.
\newblock In \emph{Proceedings of the IEEE/CVF Conference on Computer Vision and Pattern Recognition}, pages 11681--11690, 2023.

\bibitem[Dai et~al.(2023)Dai, Li, Li, Tiong, Zhao, Wang, Li, Fung, and Hoi]{dai2023instructblip}
W.~Dai, J.~Li, D.~Li, A.~Tiong, J.~Zhao, W.~Wang, B.~Li, P.~Fung, and S.~Hoi.
\newblock Instruct{BLIP}: Towards general-purpose vision-language models with instruction tuning.
\newblock In \emph{Thirty-seventh Conference on Neural Information Processing Systems}, 2023.
\newblock URL \url{https://openreview.net/forum?id=vvoWPYqZJA}.

\bibitem[Dosovitskiy et~al.(2020)Dosovitskiy, Beyer, Kolesnikov, Weissenborn, Zhai, Unterthiner, Dehghani, Minderer, Heigold, Gelly, et~al.]{dosovitskiy2020image}
A.~Dosovitskiy, L.~Beyer, A.~Kolesnikov, D.~Weissenborn, X.~Zhai, T.~Unterthiner, M.~Dehghani, M.~Minderer, G.~Heigold, S.~Gelly, et~al.
\newblock An image is worth 16x16 words: Transformers for image recognition at scale.
\newblock \emph{arXiv preprint arXiv:2010.11929}, 2020.

\bibitem[Gafni et~al.(2022)Gafni, Polyak, Ashual, Sheynin, Parikh, and Taigman]{Gafni2022Make-A-Scene}
O.~Gafni, A.~Polyak, O.~Ashual, S.~Sheynin, D.~Parikh, and Y.~Taigman.
\newblock {Make-A-Scene: Scene-Based Text-to-Image Generation with Human Priors}.
\newblock In \emph{ECCV}, 2022.
\newblock URL \url{http://arxiv.org/abs/2203.13131}.

\bibitem[Goyal et~al.(2017)Goyal, Khot, Summers-Stay, Batra, and Parikh]{goyal2017making}
Y.~Goyal, T.~Khot, D.~Summers-Stay, D.~Batra, and D.~Parikh.
\newblock Making the v in vqa matter: Elevating the role of image understanding in visual question answering.
\newblock In \emph{Proceedings of the IEEE conference on computer vision and pattern recognition}, pages 6904--6913, 2017.

\bibitem[Ho and Salimans(2021)]{ho2021classifierfree}
J.~Ho and T.~Salimans.
\newblock Classifier-free diffusion guidance.
\newblock In \emph{NeurIPS 2021 Workshop on Deep Generative Models and Downstream Applications}, 2021.
\newblock URL \url{https://openreview.net/forum?id=qw8AKxfYbI}.

\bibitem[Honnibal et~al.(2020)Honnibal, Montani, Landeghem, and Boyd]{Spacy2020}
M.~Honnibal, I.~Montani, S.~V. Landeghem, and A.~Boyd.
\newblock spacy: Industrial-strength natural language processing in python, 2020.
\newblock URL \url{https://spacy.io}.

\bibitem[Hsieh et~al.(2023)Hsieh, Zhang, Ma, Kembhavi, and Krishna]{hsieh2023sugarcrepe}
C.-Y. Hsieh, J.~Zhang, Z.~Ma, A.~Kembhavi, and R.~Krishna.
\newblock Sugarcrepe: Fixing hackable benchmarks for vision-language compositionality.
\newblock In \emph{Thirty-Seventh Conference on Neural Information Processing Systems Datasets and Benchmarks Track}, 2023.

\bibitem[Kamath et~al.(2021)Kamath, Singh, LeCun, Misra, Synnaeve, and Carion]{kamath2021mdetr}
A.~Kamath, M.~Singh, Y.~LeCun, I.~Misra, G.~Synnaeve, and N.~Carion.
\newblock Mdetr--modulated detection for end-to-end multi-modal understanding.
\newblock \emph{arXiv preprint arXiv:2104.12763}, 2021.

\bibitem[Kamath et~al.(2023)Kamath, Hessel, and Chang]{kamath2023whats}
A.~Kamath, J.~Hessel, and K.-W. Chang.
\newblock What’s “up” with vision-language models? investigating their struggle with spatial reasoning.
\newblock In \emph{Proceedings of the 2023 Conference on Empirical Methods in Natural Language Processing}, pages 9161--9175, 2023.

\bibitem[Kazemzadeh et~al.(2014)Kazemzadeh, Ordonez, Matten, and Berg]{kazemzadeh2014referitgame}
S.~Kazemzadeh, V.~Ordonez, M.~Matten, and T.~Berg.
\newblock Referitgame: Referring to objects in photographs of natural scenes.
\newblock In \emph{Proceedings of the 2014 conference on empirical methods in natural language processing (EMNLP)}, pages 787--798, 2014.

\bibitem[Khattak et~al.(2023)Khattak, Rasheed, Maaz, Khan, and Khan]{Khattak2023Maple}
M.~U. Khattak, H.~Rasheed, M.~Maaz, S.~Khan, and F.~S. Khan.
\newblock {MaPLe: Multi-modal Prompt Learning}.
\newblock In \emph{CVPR}, 2023.
\newblock ISBN 9798350301298.
\newblock \doi{10.1109/CVPR52729.2023.01832}.

\bibitem[Kirillov et~al.(2023)Kirillov, Mintun, Ravi, Mao, Rolland, Gustafson, Xiao, Whitehead, Berg, Lo, Doll{\'a}r, and Girshick]{kirillov2023segany}
A.~Kirillov, E.~Mintun, N.~Ravi, H.~Mao, C.~Rolland, L.~Gustafson, T.~Xiao, S.~Whitehead, A.~C. Berg, W.-Y. Lo, P.~Doll{\'a}r, and R.~Girshick.
\newblock Segment anything.
\newblock \emph{arXiv:2304.02643}, 2023.

\bibitem[Kornblith et~al.(2023)Kornblith, Li, Wang, and Nguyen]{Kornblith2023CFGCaption}
S.~Kornblith, L.~Li, Z.~Wang, and T.~Nguyen.
\newblock {Guiding Image Captioning Models Toward More Specific Captions}.
\newblock In \emph{ICCV}, 2023.
\newblock URL \url{http://arxiv.org/abs/2307.16686}.

\bibitem[Leng et~al.(2023)Leng, Zhang, Chen, Li, Lu, Miao, and Bing]{Leng2023HallucinationVCD}
S.~Leng, H.~Zhang, G.~Chen, X.~Li, S.~Lu, C.~Miao, and L.~Bing.
\newblock {Mitigating Object Hallucinations in Large Vision-Language Models through Visual Contrastive Decoding}, 2023.
\newblock URL \url{http://arxiv.org/abs/2311.16922}.

\bibitem[Li et~al.(2023{\natexlab{a}})Li, Li, Savarese, and Hoi]{li2023blip}
J.~Li, D.~Li, S.~Savarese, and S.~Hoi.
\newblock Blip-2: Bootstrapping language-image pre-training with frozen image encoders and large language models.
\newblock \emph{arXiv preprint arXiv:2301.12597}, 2023{\natexlab{a}}.

\bibitem[Li* et~al.(2022)Li*, Zhang*, Zhang*, Yang, Li, Zhong, Wang, Yuan, Zhang, Hwang, Chang, and Gao]{li2021grounded}
L.~H. Li*, P.~Zhang*, H.~Zhang*, J.~Yang, C.~Li, Y.~Zhong, L.~Wang, L.~Yuan, L.~Zhang, J.-N. Hwang, K.-W. Chang, and J.~Gao.
\newblock Grounded language-image pre-training.
\newblock In \emph{CVPR}, 2022.

\bibitem[Li et~al.(2023{\natexlab{b}})Li, Holtzman, Fried, Liang, Eisner, Hashimoto, Zettlemoyer, and Lewis]{li-etal-2023-contrastive}
X.~L. Li, A.~Holtzman, D.~Fried, P.~Liang, J.~Eisner, T.~Hashimoto, L.~Zettlemoyer, and M.~Lewis.
\newblock Contrastive decoding: Open-ended text generation as optimization.
\newblock In A.~Rogers, J.~Boyd-Graber, and N.~Okazaki, editors, \emph{Proceedings of the 61st Annual Meeting of the Association for Computational Linguistics (Volume 1: Long Papers)}, 2023{\natexlab{b}}.

\bibitem[Lin et~al.(2014)Lin, Maire, Belongie, Hays, Perona, Ramanan, Doll{\'a}r, and Zitnick]{lin2014microsoft}
T.-Y. Lin, M.~Maire, S.~Belongie, J.~Hays, P.~Perona, D.~Ramanan, P.~Doll{\'a}r, and C.~L. Zitnick.
\newblock Microsoft coco: Common objects in context.
\newblock In \emph{Computer Vision--ECCV 2014: 13th European Conference, Zurich, Switzerland, September 6-12, 2014, Proceedings, Part V 13}, pages 740--755. Springer, 2014.

\bibitem[Liu et~al.(2023{\natexlab{a}})Liu, Li, Wu, and Lee]{liu2023visual}
H.~Liu, C.~Li, Q.~Wu, and Y.~J. Lee.
\newblock Visual instruction tuning.
\newblock In \emph{Thirty-seventh Conference on Neural Information Processing Systems}, 2023{\natexlab{a}}.
\newblock URL \url{https://openreview.net/forum?id=w0H2xGHlkw}.

\bibitem[Liu et~al.(2024{\natexlab{a}})Liu, Li, Li, Li, Zhang, Shen, and Lee]{liu2024llavanext}
H.~Liu, C.~Li, Y.~Li, B.~Li, Y.~Zhang, S.~Shen, and Y.~J. Lee.
\newblock Llava-next: Improved reasoning, ocr, and world knowledge, January 2024{\natexlab{a}}.
\newblock URL \url{https://llava-vl.github.io/blog/2024-01-30-llava-next/}.

\bibitem[Liu et~al.(2024{\natexlab{b}})Liu, Li, Wu, and Lee]{liu2024visual}
H.~Liu, C.~Li, Q.~Wu, and Y.~J. Lee.
\newblock Visual instruction tuning.
\newblock \emph{Advances in neural information processing systems}, 36, 2024{\natexlab{b}}.

\bibitem[Liu et~al.(2023{\natexlab{b}})Liu, Zeng, Ren, Li, Zhang, Yang, Li, Yang, Su, Zhu, et~al.]{liu2023grounding}
S.~Liu, Z.~Zeng, T.~Ren, F.~Li, H.~Zhang, J.~Yang, C.~Li, J.~Yang, H.~Su, J.~Zhu, et~al.
\newblock Grounding dino: Marrying dino with grounded pre-training for open-set object detection.
\newblock \emph{arXiv preprint arXiv:2303.05499}, 2023{\natexlab{b}}.

\bibitem[Liu et~al.(2022)Liu, Guo, Yin, Song, Liu, Nie, and Zhang]{liu2022answer}
Y.~Liu, Y.~Guo, J.~Yin, X.~Song, W.~Liu, L.~Nie, and M.~Zhang.
\newblock Answer questions with right image regions: A visual attention regularization approach.
\newblock \emph{ACM Transactions on Multimedia Computing, Communications, and Applications (TOMM)}, 18\penalty0 (4):\penalty0 1--18, 2022.

\bibitem[Ma et~al.(2023)Ma, Hong, Gul, Gandhi, Gao, and Krishna]{ma2023crepe}
Z.~Ma, J.~Hong, M.~O. Gul, M.~Gandhi, I.~Gao, and R.~Krishna.
\newblock Crepe: Can vision-language foundation models reason compositionally?
\newblock In \emph{Proceedings of the IEEE/CVF Conference on Computer Vision and Pattern Recognition}, pages 10910--10921, 2023.

\bibitem[Mao et~al.(2015)Mao, Huang, Toshev, Camburu, Yuille, and Murphy]{Mao2015GenerationAC}
J.~Mao, J.~Huang, A.~Toshev, O.-M. Camburu, A.~L. Yuille, and K.~P. Murphy.
\newblock Generation and comprehension of unambiguous object descriptions.
\newblock \emph{2016 IEEE Conference on Computer Vision and Pattern Recognition (CVPR)}, pages 11--20, 2015.
\newblock URL \url{https://api.semanticscholar.org/CorpusID:8745888}.

\bibitem[O'Brien and Lewis(2023)]{obrien2023contrastive}
S.~O'Brien and M.~Lewis.
\newblock Contrastive decoding improves reasoning in large language models, 2023.

\bibitem[OpenAI et~al.(2023)OpenAI, :, Achiam, Adler, Agarwal, Ahmad, Akkaya, Aleman, Almeida, Altenschmidt, Altman, Anadkat, Avila, Babuschkin, Balaji, Balcom, Baltescu, Bao, Bavarian, Belgum, Bello, Berdine, Bernadett-Shapiro, Berner, Bogdonoff, Boiko, Boyd, Brakman, Brockman, Brooks, Brundage, Button, Cai, Campbell, Cann, Carey, Carlson, Carmichael, Chan, Chang, Chantzis, Chen, Chen, Chen, Chen, Chen, Chess, Cho, Chu, Chung, Cummings, Currier, Dai, Decareaux, Degry, Deutsch, Deville, Dhar, Dohan, Dowling, Dunning, Ecoffet, Eleti, Eloundou, Farhi, Fedus, Felix, Fishman, Forte, Fulford, Gao, Georges, Gibson, Goel, Gogineni, Goh, Gontijo-Lopes, Gordon, Grafstein, Gray, Greene, Gross, Gu, Guo, Hallacy, Han, Harris, He, Heaton, Heidecke, Hesse, Hickey, Hickey, Hoeschele, Houghton, Hsu, Hu, Hu, Huizinga, Jain, Jain, Jang, Jiang, Jiang, Jin, Jin, Jomoto, Jonn, Jun, Kaftan, Łukasz Kaiser, Kamali, Kanitscheider, Keskar, Khan, Kilpatrick, Kim, Kim, Kim, Kirchner, Kiros, Knight, Kokotajlo, Łukasz Kondraciuk,
  Kondrich, Konstantinidis, Kosic, Krueger, Kuo, Lampe, Lan, Lee, Leike, Leung, Levy, Li, Lim, Lin, Lin, Litwin, Lopez, Lowe, Lue, Makanju, Malfacini, Manning, Markov, Markovski, Martin, Mayer, Mayne, McGrew, McKinney, McLeavey, McMillan, McNeil, Medina, Mehta, Menick, Metz, Mishchenko, Mishkin, Monaco, Morikawa, Mossing, Mu, Murati, Murk, Mély, Nair, Nakano, Nayak, Neelakantan, Ngo, Noh, Ouyang, O'Keefe, Pachocki, Paino, Palermo, Pantuliano, Parascandolo, Parish, Parparita, Passos, Pavlov, Peng, Perelman, de~Avila Belbute~Peres, Petrov, de~Oliveira~Pinto, Michael, Pokorny, Pokrass, Pong, Powell, Power, Power, Proehl, Puri, Radford, Rae, Ramesh, Raymond, Real, Rimbach, Ross, Rotsted, Roussez, Ryder, Saltarelli, Sanders, Santurkar, Sastry, Schmidt, Schnurr, Schulman, Selsam, Sheppard, Sherbakov, Shieh, Shoker, Shyam, Sidor, Sigler, Simens, Sitkin, Slama, Sohl, Sokolowsky, Song, Staudacher, Such, Summers, Sutskever, Tang, Tezak, Thompson, Tillet, Tootoonchian, Tseng, Tuggle, Turley, Tworek, Uribe, Vallone,
  Vijayvergiya, Voss, Wainwright, Wang, Wang, Wang, Ward, Wei, Weinmann, Welihinda, Welinder, Weng, Weng, Wiethoff, Willner, Winter, Wolrich, Wong, Workman, Wu, Wu, Wu, Xiao, Xu, Yoo, Yu, Yuan, Zaremba, Zellers, Zhang, Zhang, Zhao, Zheng, Zhuang, Zhuk, and Zoph]{openai2023gpt4}
OpenAI, :, J.~Achiam, S.~Adler, S.~Agarwal, L.~Ahmad, I.~Akkaya, F.~L. Aleman, D.~Almeida, J.~Altenschmidt, S.~Altman, S.~Anadkat, R.~Avila, I.~Babuschkin, S.~Balaji, V.~Balcom, P.~Baltescu, H.~Bao, M.~Bavarian, J.~Belgum, I.~Bello, J.~Berdine, G.~Bernadett-Shapiro, C.~Berner, L.~Bogdonoff, O.~Boiko, M.~Boyd, A.-L. Brakman, G.~Brockman, T.~Brooks, M.~Brundage, K.~Button, T.~Cai, R.~Campbell, A.~Cann, B.~Carey, C.~Carlson, R.~Carmichael, B.~Chan, C.~Chang, F.~Chantzis, D.~Chen, S.~Chen, R.~Chen, J.~Chen, M.~Chen, B.~Chess, C.~Cho, C.~Chu, H.~W. Chung, D.~Cummings, J.~Currier, Y.~Dai, C.~Decareaux, T.~Degry, N.~Deutsch, D.~Deville, A.~Dhar, D.~Dohan, S.~Dowling, S.~Dunning, A.~Ecoffet, A.~Eleti, T.~Eloundou, D.~Farhi, L.~Fedus, N.~Felix, S.~P. Fishman, J.~Forte, I.~Fulford, L.~Gao, E.~Georges, C.~Gibson, V.~Goel, T.~Gogineni, G.~Goh, R.~Gontijo-Lopes, J.~Gordon, M.~Grafstein, S.~Gray, R.~Greene, J.~Gross, S.~S. Gu, Y.~Guo, C.~Hallacy, J.~Han, J.~Harris, Y.~He, M.~Heaton, J.~Heidecke, C.~Hesse, A.~Hickey,
  W.~Hickey, P.~Hoeschele, B.~Houghton, K.~Hsu, S.~Hu, X.~Hu, J.~Huizinga, S.~Jain, S.~Jain, J.~Jang, A.~Jiang, R.~Jiang, H.~Jin, D.~Jin, S.~Jomoto, B.~Jonn, H.~Jun, T.~Kaftan, Łukasz Kaiser, A.~Kamali, I.~Kanitscheider, N.~S. Keskar, T.~Khan, L.~Kilpatrick, J.~W. Kim, C.~Kim, Y.~Kim, H.~Kirchner, J.~Kiros, M.~Knight, D.~Kokotajlo, Łukasz Kondraciuk, A.~Kondrich, A.~Konstantinidis, K.~Kosic, G.~Krueger, V.~Kuo, M.~Lampe, I.~Lan, T.~Lee, J.~Leike, J.~Leung, D.~Levy, C.~M. Li, R.~Lim, M.~Lin, S.~Lin, M.~Litwin, T.~Lopez, R.~Lowe, P.~Lue, A.~Makanju, K.~Malfacini, S.~Manning, T.~Markov, Y.~Markovski, B.~Martin, K.~Mayer, A.~Mayne, B.~McGrew, S.~M. McKinney, C.~McLeavey, P.~McMillan, J.~McNeil, D.~Medina, A.~Mehta, J.~Menick, L.~Metz, A.~Mishchenko, P.~Mishkin, V.~Monaco, E.~Morikawa, D.~Mossing, T.~Mu, M.~Murati, O.~Murk, D.~Mély, A.~Nair, R.~Nakano, R.~Nayak, A.~Neelakantan, R.~Ngo, H.~Noh, L.~Ouyang, C.~O'Keefe, J.~Pachocki, A.~Paino, J.~Palermo, A.~Pantuliano, G.~Parascandolo, J.~Parish, E.~Parparita,
  A.~Passos, M.~Pavlov, A.~Peng, A.~Perelman, F.~de~Avila Belbute~Peres, M.~Petrov, H.~P. de~Oliveira~Pinto, Michael, Pokorny, M.~Pokrass, V.~Pong, T.~Powell, A.~Power, B.~Power, E.~Proehl, R.~Puri, A.~Radford, J.~Rae, A.~Ramesh, C.~Raymond, F.~Real, K.~Rimbach, C.~Ross, B.~Rotsted, H.~Roussez, N.~Ryder, M.~Saltarelli, T.~Sanders, S.~Santurkar, G.~Sastry, H.~Schmidt, D.~Schnurr, J.~Schulman, D.~Selsam, K.~Sheppard, T.~Sherbakov, J.~Shieh, S.~Shoker, P.~Shyam, S.~Sidor, E.~Sigler, M.~Simens, J.~Sitkin, K.~Slama, I.~Sohl, B.~Sokolowsky, Y.~Song, N.~Staudacher, F.~P. Such, N.~Summers, I.~Sutskever, J.~Tang, N.~Tezak, M.~Thompson, P.~Tillet, A.~Tootoonchian, E.~Tseng, P.~Tuggle, N.~Turley, J.~Tworek, J.~F.~C. Uribe, A.~Vallone, A.~Vijayvergiya, C.~Voss, C.~Wainwright, J.~J. Wang, A.~Wang, B.~Wang, J.~Ward, J.~Wei, C.~Weinmann, A.~Welihinda, P.~Welinder, J.~Weng, L.~Weng, M.~Wiethoff, D.~Willner, C.~Winter, S.~Wolrich, H.~Wong, L.~Workman, S.~Wu, J.~Wu, M.~Wu, K.~Xiao, T.~Xu, S.~Yoo, K.~Yu, Q.~Yuan, W.~Zaremba,
  R.~Zellers, C.~Zhang, M.~Zhang, S.~Zhao, T.~Zheng, J.~Zhuang, W.~Zhuk, and B.~Zoph.
\newblock Gpt-4 technical report, 2023.

\bibitem[Plummer et~al.(2017)Plummer, Wang, Cervantes, Caicedo, Hockenmaier, and Lazebnik]{flickrentitiesijcv}
B.~A. Plummer, L.~Wang, C.~M. Cervantes, J.~C. Caicedo, J.~Hockenmaier, and S.~Lazebnik.
\newblock Flickr30k entities: Collecting region-to-phrase correspondences for richer image-to-sentence models.
\newblock \emph{IJCV}, 123\penalty0 (1):\penalty0 74--93, 2017.

\bibitem[Radford et~al.(2021)Radford, Kim, Hallacy, Ramesh, Goh, Agarwal, Sastry, Askell, Mishkin, Clark, Krueger, and Sutskever]{DBLP:conf/icml/RadfordKHRGASAM21}
A.~Radford, J.~W. Kim, C.~Hallacy, A.~Ramesh, G.~Goh, S.~Agarwal, G.~Sastry, A.~Askell, P.~Mishkin, J.~Clark, G.~Krueger, and I.~Sutskever.
\newblock Learning transferable visual models from natural language supervision.
\newblock In M.~Meila and T.~Zhang, editors, \emph{Proceedings of the 38th International Conference on Machine Learning, {ICML} 2021, 18-24 July 2021, Virtual Event}, volume 139 of \emph{Proceedings of Machine Learning Research}, pages 8748--8763. {PMLR}, 2021.
\newblock URL \url{http://proceedings.mlr.press/v139/radford21a.html}.

\bibitem[Ray et~al.(2024)Ray, Radenovic, Dubey, Plummer, Krishna, and Saenko]{ray2024cola}
A.~Ray, F.~Radenovic, A.~Dubey, B.~Plummer, R.~Krishna, and K.~Saenko.
\newblock cola: A benchmark for compositional text-to-image retrieval.
\newblock \emph{Advances in Neural Information Processing Systems}, 36, 2024.

\bibitem[Ren et~al.(2024)Ren, Liu, Zeng, Lin, Li, Cao, Chen, Huang, Chen, Yan, Zeng, Zhang, Li, Yang, Li, Jiang, and Zhang]{ren2024grounded}
T.~Ren, S.~Liu, A.~Zeng, J.~Lin, K.~Li, H.~Cao, J.~Chen, X.~Huang, Y.~Chen, F.~Yan, Z.~Zeng, H.~Zhang, F.~Li, J.~Yang, H.~Li, Q.~Jiang, and L.~Zhang.
\newblock Grounded sam: Assembling open-world models for diverse visual tasks, 2024.

\bibitem[Ribeiro et~al.(2016)Ribeiro, Singh, and Guestrin]{ribeiro-etal-2016-trust}
M.~Ribeiro, S.~Singh, and C.~Guestrin.
\newblock {``}why should {I} trust you?{''}: Explaining the predictions of any classifier.
\newblock In J.~DeNero, M.~Finlayson, and S.~Reddy, editors, \emph{Proceedings of the 2016 Conference of the North {A}merican Chapter of the Association for Computational Linguistics: Demonstrations}, pages 97--101, San Diego, California, June 2016. Association for Computational Linguistics.
\newblock \doi{10.18653/v1/N16-3020}.
\newblock URL \url{https://aclanthology.org/N16-3020}.

\bibitem[Saharia et~al.(2022)Saharia, Chan, Saxena, Li, Whang, Denton, Ghasemipour, Gontijo-Lopes, Ayan, Salimans, Ho, Fleet, and Norouzi]{saharia2022photorealistic}
C.~Saharia, W.~Chan, S.~Saxena, L.~Li, J.~Whang, E.~Denton, S.~K.~S. Ghasemipour, R.~Gontijo-Lopes, B.~K. Ayan, T.~Salimans, J.~Ho, D.~J. Fleet, and M.~Norouzi.
\newblock Photorealistic text-to-image diffusion models with deep language understanding.
\newblock In A.~H. Oh, A.~Agarwal, D.~Belgrave, and K.~Cho, editors, \emph{Advances in Neural Information Processing Systems}, 2022.
\newblock URL \url{https://openreview.net/forum?id=08Yk-n5l2Al}.

\bibitem[Sanchez et~al.(2023)Sanchez, Fan, Spangher, Levi, Ammanamanchi, and Biderman]{sanchez2023stay}
G.~Sanchez, H.~Fan, A.~Spangher, E.~Levi, P.~S. Ammanamanchi, and S.~Biderman.
\newblock Stay on topic with classifier-free guidance.
\newblock \emph{arXiv preprint arXiv:2306.17806}, 2023.

\bibitem[Selvaraju et~al.(2019)Selvaraju, Lee, Shen, Jin, Ghosh, Heck, Batra, and Parikh]{selvaraju2019taking}
R.~R. Selvaraju, S.~Lee, Y.~Shen, H.~Jin, S.~Ghosh, L.~Heck, D.~Batra, and D.~Parikh.
\newblock Taking a hint: Leveraging explanations to make vision and language models more grounded.
\newblock In \emph{Proceedings of the IEEE/CVF international conference on computer vision}, pages 2591--2600, 2019.

\bibitem[Shi et~al.(2023)Shi, Han, Lewis, Tsvetkov, Zettlemoyer, and tau Yih]{shi2023trusting}
W.~Shi, X.~Han, M.~Lewis, Y.~Tsvetkov, L.~Zettlemoyer, and S.~W. tau Yih.
\newblock Trusting your evidence: Hallucinate less with context-aware decoding, 2023.

\bibitem[Shtedritski et~al.(2023)Shtedritski, Rupprecht, and Vedaldi]{Shtedritski2023}
A.~Shtedritski, C.~Rupprecht, and A.~Vedaldi.
\newblock What does clip know about a red circle? visual prompt engineering for vlms.
\newblock In \emph{Proceedings of the IEEE/CVF International Conference on Computer Vision (ICCV)}, pages 11987--11997, October 2023.

\bibitem[Singh et~al.(2022)Singh, Hu, Goswami, Couairon, Galuba, Rohrbach, and Kiela]{singh2022flava}
A.~Singh, R.~Hu, V.~Goswami, G.~Couairon, W.~Galuba, M.~Rohrbach, and D.~Kiela.
\newblock {FLAVA:} {A} foundational language and vision alignment model.
\newblock In \emph{CVPR}, 2022.

\bibitem[Sun et~al.(2023)Sun, Fang, Wu, Zhang, Zang, Kong, Xiong, Lin, and Wang]{sun2023alphaclip}
Z.~Sun, Y.~Fang, T.~Wu, P.~Zhang, Y.~Zang, S.~Kong, Y.~Xiong, D.~Lin, and J.~Wang.
\newblock Alpha-clip: A clip model focusing on wherever you want, 2023.

\bibitem[Thrush et~al.(2022)Thrush, Jiang, Bartolo, Singh, Williams, Kiela, and Ross]{thrush2022winoground}
T.~Thrush, R.~Jiang, M.~Bartolo, A.~Singh, A.~Williams, D.~Kiela, and C.~Ross.
\newblock Winoground: Probing vision and language models for visio-linguistic compositionality.
\newblock In \emph{Proceedings of the IEEE/CVF Conference on Computer Vision and Pattern Recognition}, pages 5238--5248, 2022.

\bibitem[Wang et~al.(2023)Wang, Saharia, Montgomery, Pont-Tuset, Noy, Pellegrini, Onoe, Laszlo, Fleet, Soricut, Baldridge, Norouzi, Anderson, and Chan]{10204528}
S.~Wang, C.~Saharia, C.~Montgomery, J.~Pont-Tuset, S.~Noy, S.~Pellegrini, Y.~Onoe, S.~Laszlo, D.~J. Fleet, R.~Soricut, J.~Baldridge, M.~Norouzi, P.~Anderson, and W.~Chan.
\newblock Imagen editor and editbench: Advancing and evaluating text-guided image inpainting.
\newblock In \emph{2023 IEEE/CVF Conference on Computer Vision and Pattern Recognition (CVPR)}, pages 18359--18369, Los Alamitos, CA, USA, jun 2023. IEEE Computer Society.
\newblock \doi{10.1109/CVPR52729.2023.01761}.
\newblock URL \url{https://doi.ieeecomputersociety.org/10.1109/CVPR52729.2023.01761}.

\bibitem[Wu and Mooney(2019)]{wu2019self}
J.~Wu and R.~Mooney.
\newblock Self-critical reasoning for robust visual question answering.
\newblock \emph{Advances in Neural Information Processing Systems}, 32, 2019.

\bibitem[Yang et~al.(2023)Yang, Zhang, Li, Zou, Li, and Gao]{Yang2023SoM}
J.~Yang, H.~Zhang, F.~Li, X.~Zou, C.~Li, and J.~Gao.
\newblock {Set-of-Mark Prompting Unleashes Extraordinary Visual Grounding in GPT-4V}, 2023.
\newblock URL \url{http://arxiv.org/abs/2310.11441}.

\bibitem[Yao et~al.(2021)Yao, Zhang, Zhang, Liu, Chua, and Sun]{Yao2021CPT}
Y.~Yao, A.~Zhang, Z.~Zhang, Z.~Liu, T.-S. Chua, and M.~Sun.
\newblock {CPT: Colorful Prompt Tuning for Pre-trained Vision-Language Models}, 2021.
\newblock URL \url{http://arxiv.org/abs/2109.11797}.

\bibitem[Yarom et~al.(2023)Yarom, Bitton, Changpinyo, Aharoni, Herzig, Lang, Ofek, and Szpektor]{yarom2023what}
M.~Yarom, Y.~Bitton, S.~Changpinyo, R.~Aharoni, J.~Herzig, O.~Lang, E.~Ofek, and I.~Szpektor.
\newblock What you see is what you read? improving text-image alignment evaluation.
\newblock In \emph{Thirty-seventh Conference on Neural Information Processing Systems}, 2023.
\newblock URL \url{https://openreview.net/forum?id=j5AoleAIru}.

\bibitem[Ying et~al.(2022)Ying, Hase, and Bansal]{ying2022visfis}
Z.~Ying, P.~Hase, and M.~Bansal.
\newblock Visfis: Visual feature importance supervision with right-for-the-right-reason objectives.
\newblock \emph{Advances in Neural Information Processing Systems}, 35:\penalty0 17057--17072, 2022.

\bibitem[Young et~al.(2014)Young, Lai, Hodosh, and Hockenmaier]{young2014image}
P.~Young, A.~Lai, M.~Hodosh, and J.~Hockenmaier.
\newblock From image descriptions to visual denotations: New similarity metrics for semantic inference over event descriptions.
\newblock \emph{Transactions of the Association for Computational Linguistics}, 2:\penalty0 67--78, 2014.

\bibitem[Zellers et~al.(2021)Zellers, Lu, Hessel, Yu, Park, Cao, Farhadi, and Choi]{Zellers2021Merlot}
R.~Zellers, X.~Lu, J.~Hessel, Y.~Yu, J.~S. Park, J.~Cao, A.~Farhadi, and Y.~Choi.
\newblock {MERLOT: Multimodal Neural Script Knowledge Models}.
\newblock In \emph{NeurIPS}, 2021.
\newblock URL \url{http://arxiv.org/abs/2106.02636}.

\bibitem[Zhang et~al.(2022)Zhang, Zhang, Hu, Chen, Li, Dai, Wang, Yuan, Hwang, and Gao]{zhang2022glipv2}
H.~Zhang, P.~Zhang, X.~Hu, Y.-C. Chen, L.~H. Li, X.~Dai, L.~Wang, L.~Yuan, J.-N. Hwang, and J.~Gao.
\newblock Glipv2: Unifying localization and vision-language understanding.
\newblock \emph{arXiv preprint arXiv:2206.05836}, 2022.

\bibitem[Zhang et~al.(2016)Zhang, Goyal, Summers-Stay, Batra, and Parikh]{zhang2016yin}
P.~Zhang, Y.~Goyal, D.~Summers-Stay, D.~Batra, and D.~Parikh.
\newblock Yin and yang: Balancing and answering binary visual questions.
\newblock In \emph{Proceedings of the IEEE conference on computer vision and pattern recognition}, pages 5014--5022, 2016.

\bibitem[Zhang et~al.(2023)Zhang, Sun, Chen, Xiao, Shao, Zhang, Liu, Chen, and Luo]{zhang2023gpt4roi}
S.~Zhang, P.~Sun, S.~Chen, M.~Xiao, W.~Shao, W.~Zhang, Y.~Liu, K.~Chen, and P.~Luo.
\newblock Gpt4roi: Instruction tuning large language model on region-of-interest, 2023.

\bibitem[Zhao et~al.(2024)Zhao, Deng, Zhang, and Gu]{zhao2024mitigating}
L.~Zhao, Y.~Deng, W.~Zhang, and Q.~Gu.
\newblock Mitigating object hallucination in large vision-language models via classifier-free guidance, 2024.

\bibitem[Zou et~al.(2023)Zou, Yang, Zhang, Li, Li, Wang, Wang, Gao, and Lee]{zou2023segment}
X.~Zou, J.~Yang, H.~Zhang, F.~Li, L.~Li, J.~Wang, L.~Wang, J.~Gao, and Y.~J. Lee.
\newblock Segment everything everywhere all at once.
\newblock In \emph{Thirty-seventh Conference on Neural Information Processing Systems}, 2023.
\newblock URL \url{https://openreview.net/forum?id=UHBrWeFWlL}.

\end{thebibliography}

\appendix

\vspace{20pt}

{
\Large
\textbf{{Appendix}}
}

In this appendix, we include details of the experiment setup (\cref{sec:experimental_setup_appendix}) and qualitative examples (\cref{sec:qualitative_examples}).

\section{Experimental Setup Details}
\label{sec:experimental_setup_appendix}

\subsection{Dataset Statistics}
In \cref{tab:dataset_statistics}, we show the statistics of the datasets used in our experiments.

\begin{table}[th]
    \caption{Dataset Statistics. \textsc{REC} denotes Referring Expression Comprehension. 
    }
    \label{tab:dataset_statistics}
    \centering
    \begin{tabular}{l c}
    \toprule
    Dataset & \# examples \\
    \midrule
    ViP-Bench & 303\\
    \midrule
    \multicolumn{2}{c}{\textsc{Image-Text Alignment}}\\
    \midrule
    \whatsup{} & 3,280 \\
    \sugarcrepe{} swap-att & 1,332 \\
    \sugarcrepe{} swap-obj & 490 \\
    \seetrue{} DrawBench & 1,312 \\
    \seetrue{} EditBench & 3,827 \\
    \seetrue{} COCO-t2i & 1,791 \\
    \midrule
    \multicolumn{2}{c}{\textsc{REC and Phrase Grounding}} \\
    \midrule
    RefCOCO testA & 5,657 \\
    RefCOCO testB & 5,095 \\
    RefCOCO+ testA & 5,726 \\
    RefCOCO+ testB & 4,889 \\
    RefCOCOg test & 9,602 \\
    Flickr30K Entities test & 4,969 \\
    \bottomrule
    \end{tabular}
\end{table}

\subsection{Details on \METHOD{} Prompting, Region Proposals, and Probability Extraction}
To generate answers for VQA-style tasks,
such as ViP-Bench (\cref{sec:vip-bench}) and \whatsup{} (\cref{sec:whatsup}), we directly provide the model with the question to generate the response. Following \cite{kamath2023whats}, we use the probability of \texttt{Yes} as the score of each image-text pair for \whatsup{}.
For other tasks, we use the prompt \texttt{Provide a one-sentence caption for the provided image}, which has been seen by the models tested during the pre-training phase \cite{liu2023visual,liu2024llavanext}. Then, we force the model to decode the sentence or phrase to retrieve the probabilities of the tokens. 

For the bounding box extractions on Referring Expression Comprehension (REC), we use \textit{positive tokens} from the pre-processed data provided by Kamath~\etal{}~\cite{kamath2021mdetr} for generating bounding box candidates. For phrase grounding, we directly use the phrase to extract bounding box proposals.

\subsection{Details on Set-of-Mark Prompting}\label{sec:som_appendix}

To implement SoM prompting, we transform any mentions of \texttt{``within the \{color\} rectangle''} to \texttt{``in \{number\}''} in the text prompt via regular expression. 
For example, the question "Are the numbers \textbf{within the red rectangle} and \textbf{within the purple rectangle} the same?" becomes "Are the numbers in \textbf{0} and in \textbf{1} the same?", where the image for SoM removes the bounding box and instead contains an overlay on the object with numbering. 
We apply the inverse process to the outputs, transforming mentions of the numbers back to bounding boxes with colors in the answer, to be compatible with the original scoring methods. 
For example, the answer ``Yes, the numbers in \textbf{0} and in \textbf{1} are the same." is transformed into ``Yes, the numbers \textbf{within the red rectangle} and \textbf{within the purple rectangle} are the same."

\begin{figure}[ht]
    \centering
    \begin{tabularx}{.8\linewidth}{ X }
    \toprule
    \textbf{Question}: What is the value of the variable in the equation within the purple rectangle? \\
    \begin{center}
        \includegraphics[width=0.35\linewidth]{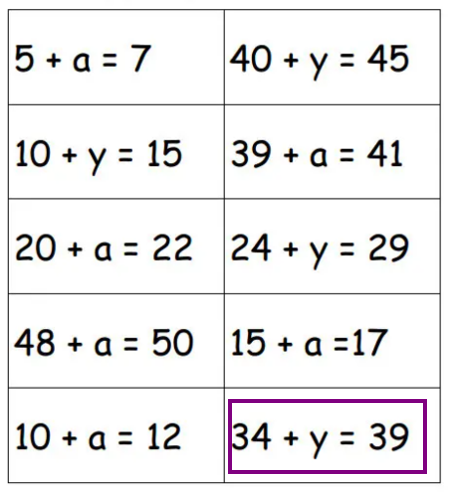}
    \end{center}\\
    \midrule
    \textbf{\vipllavamodel{}}: The variable in the equation within the purple rectangle is 34. \\
    \textbf{\llavalarge{}}: The equation within the purple rectangle is: 10 + a = 12. To find the value of the variable `a', you need to solve for `a'. You can do this by subtracting 10 from both sides of the equation: a = 12 - 10. a = 2. So the value of `a' in the equation within the purple rectangle is 2. \\
    \textbf{\METHOD{} (Ours)}: The equation within the purple rectangle reads: 34 + y = 39. To find the value of the variable `y', you subtract 34 from both sides of the equation: 34 + y = 39. y = 39 - 34. y = 5. So, the value of `y' in this equation is 5. \\
    \bottomrule
    \end{tabularx}
    \caption{An example from ViP-Bench for assessing \math{} and \ocr{} capabilities.}
    \label{fig:qualitative_examples_1}
\end{figure}

\begin{figure}[t!]
    \centering
    \begin{tabularx}{\linewidth}{ X }
    \toprule
    \textbf{Question}: What is the color of the clothing of the person within the yellow rectangle? \\
    \begin{center}
        \includegraphics[width=0.95\linewidth]{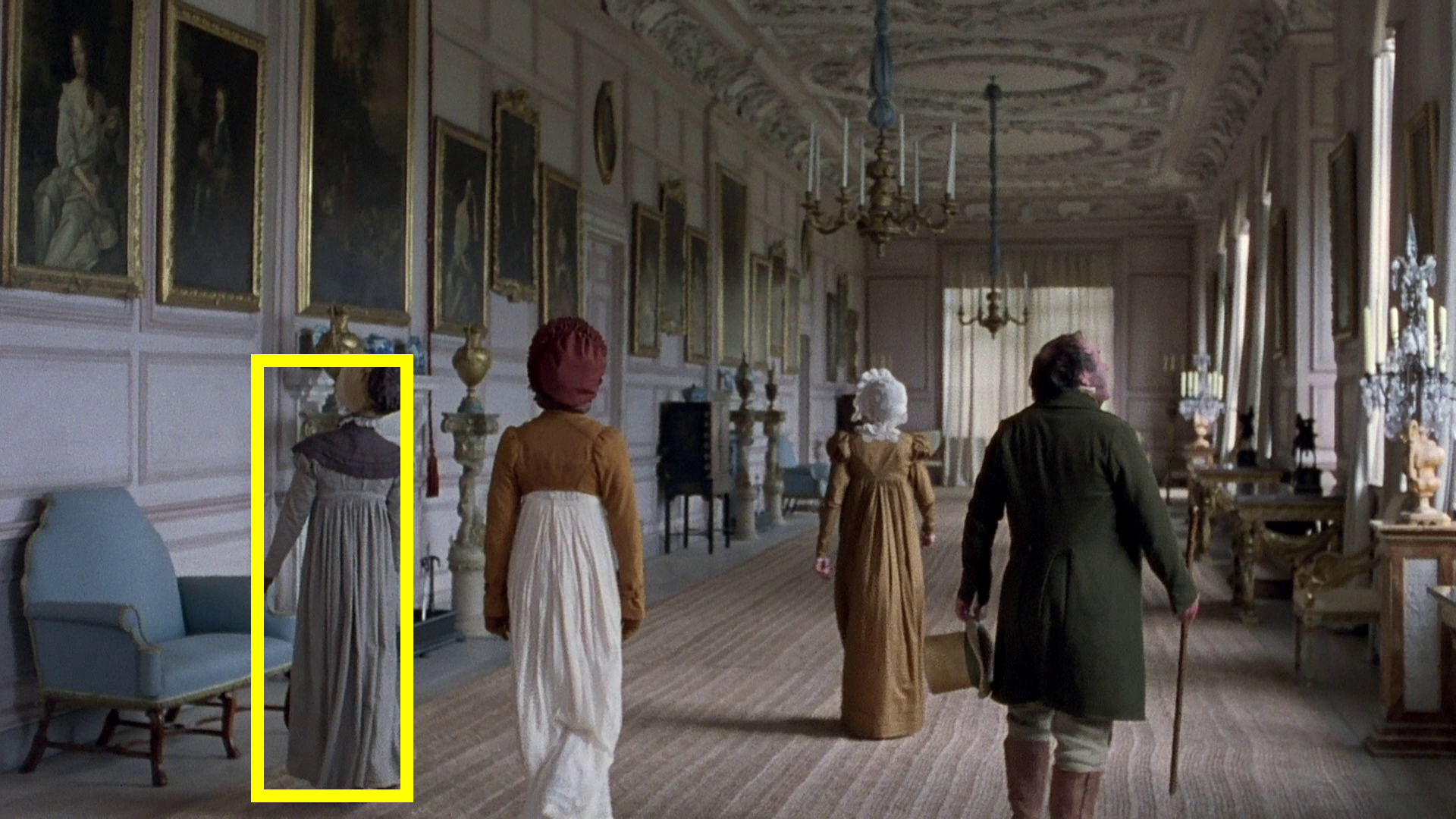}
    \end{center}\\
    \midrule
    \textbf{\vipllava{}}: The person within the yellow rectangle is wearing white clothing. \\
    \textbf{\llavalarge{}}: The person within the yellow rectangle is wearing a brown coat and a white shirt.\\
    \textbf{\METHOD{} (Ours)}: The person within the yellow rectangle is wearing a long, gray dress with what appears to be a fitted bodice and a flowing skirt.\\
    \bottomrule
    \end{tabularx}
    \caption{An example from ViP-Bench for assessing \rec{} capability.}
    \label{fig:qualitative_examples_2}
\end{figure}

\begin{figure}[t!]
    \centering
    \begin{tabularx}{\linewidth}{ X }
    \toprule
    \textbf{Question}: Question: Which plant is the tallest? Plant 1: the plant within the purple rectangle, Plant 2: the plant within the red rectangle, or Plant 3: the plant within the yellow rectangle? \\
    \begin{center}
        \includegraphics[width=\linewidth]{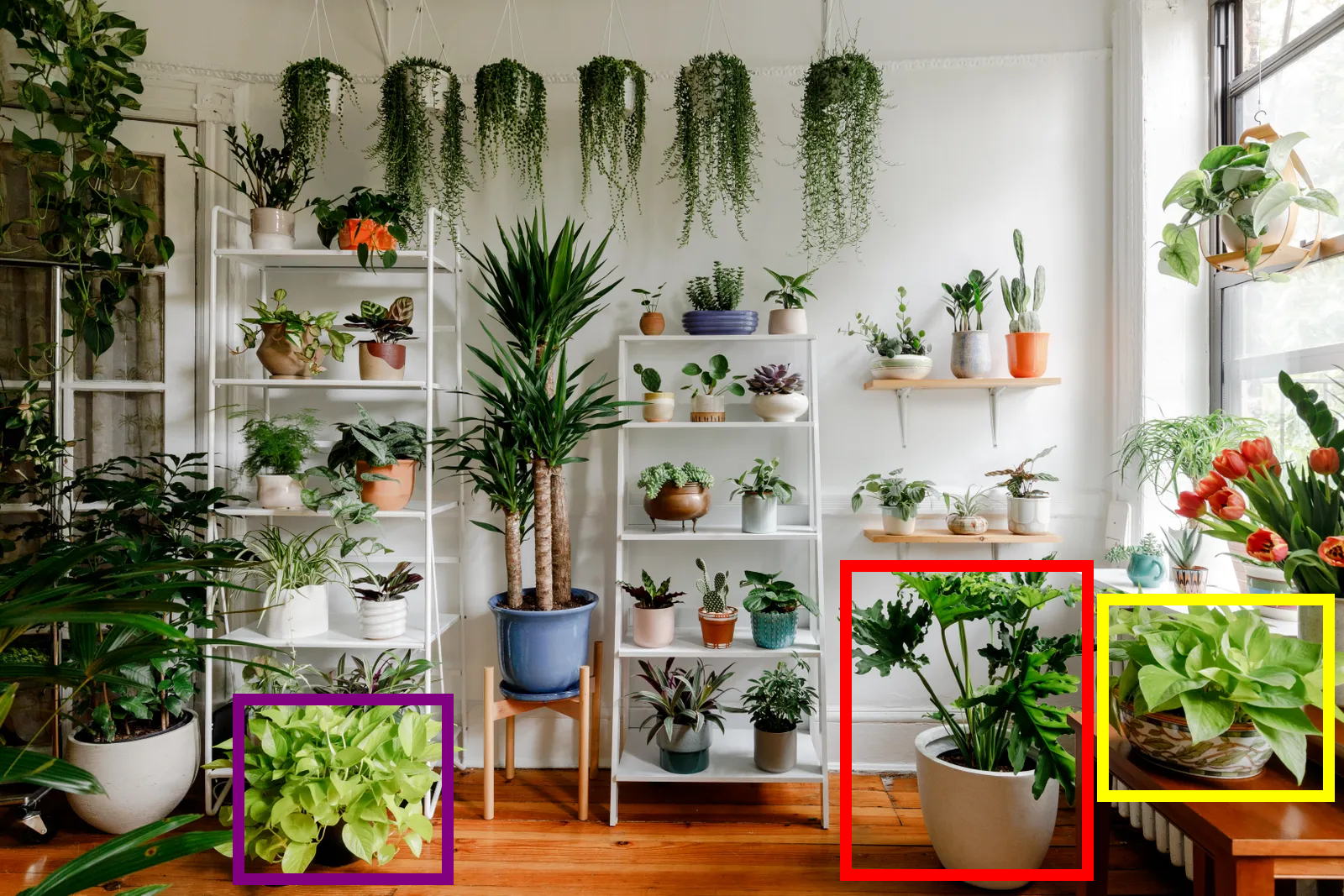}
    \end{center}\\
    \midrule
    \textbf{\vipllavamodel{}}: Plant 1, the plant within the purple rectangle, is the tallest. \\
    \textbf{\llavalarge{}}: The tallest plant in the image is the one within the purple rectangle. \\
    \textbf{\METHOD{} (Ours)}: Based on the image provided, Plant 2 within the red rectangle appears to be the tallest among the three highlighted plants. \\
    \bottomrule
    \end{tabularx}
    \caption{An example from ViP-Bench for assessing \rec{} capability.}
    \label{fig:qualitative_examples_3}
\end{figure}

\begin{figure}[t!]
    \centering
    \begin{tabularx}{\linewidth}{ X }
    \toprule
    \textbf{Question}: What is the color of the clothing of the person within the yellow rectangle? \\
    \begin{center}
        \includegraphics[width=\linewidth]{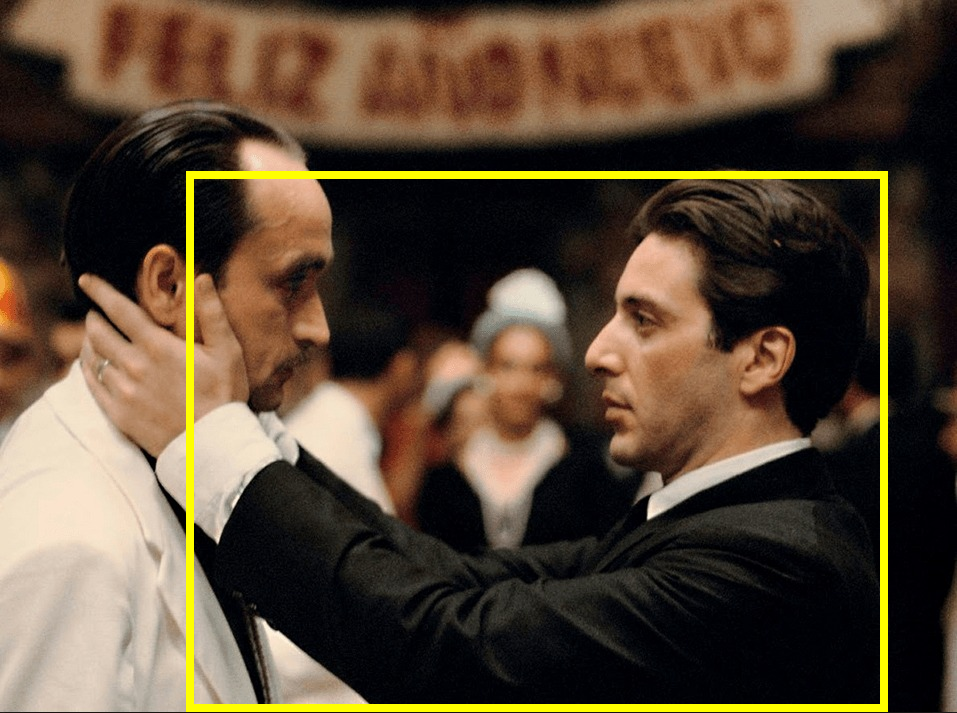}
    \end{center}\\
    \midrule
    \textbf{\vipllavamodel{}}: The person within the yellow rectangle is wearing a black suit. \\
    \textbf{\llavalarge{}}: The person within the yellow rectangle is wearing a white shirt. \\
    \textbf{\METHOD{} (Ours)}: The person within the yellow rectangle is wearing black clothing. \\
    \bottomrule
    \end{tabularx}
    \caption{An example from ViP-Bench for assessing \rec{} and \know{} capabilities.}
    \label{fig:qualitative_examples_4}
\end{figure}

\begin{figure}[t!]
    \centering
    \begin{tabularx}{\linewidth}{ X }
    \toprule
    \textbf{Text}: A \textbf{car} on a \textbf{road} passes a standing \textbf{elephant}. \\
    \begin{center}
        \includegraphics[width=0.75\linewidth]{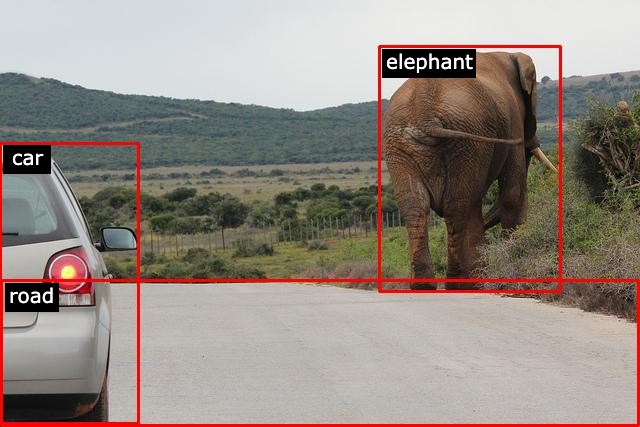}
    \end{center}\\
    \midrule
    \textbf{Text}: The extremely small \textbf{car} is parked behind the \textbf{bus}.\\
    \begin{center}
        \includegraphics[width=0.75\linewidth]{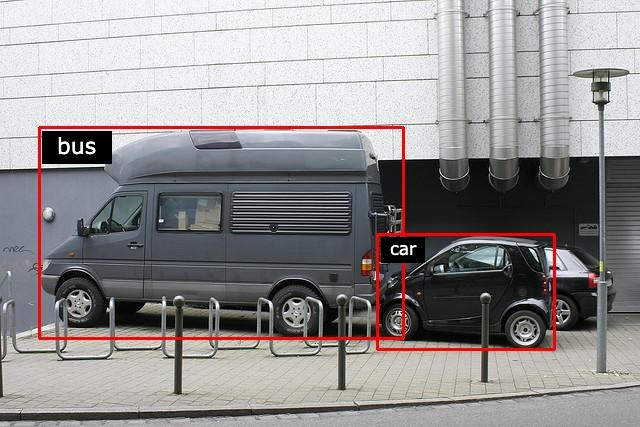}
    \end{center}\\
    \bottomrule
    \end{tabularx}
    \caption{Qualitative examples of selected regions for \sugarcrepe{}. We bold the noun phrases used to propose the regions and overlay the reranked regions for illustrative purposes.
    }
    \label{fig:qualitative_examples_bbox_sugarcrepe}
\end{figure}

\begin{figure}[t!]
    \centering
    \begin{tabularx}{\linewidth}{ X }
    \toprule
    \textbf{Text}: \textbf{cake} with cherries on top. \\
    \begin{center}
        \includegraphics[width=0.75\linewidth]{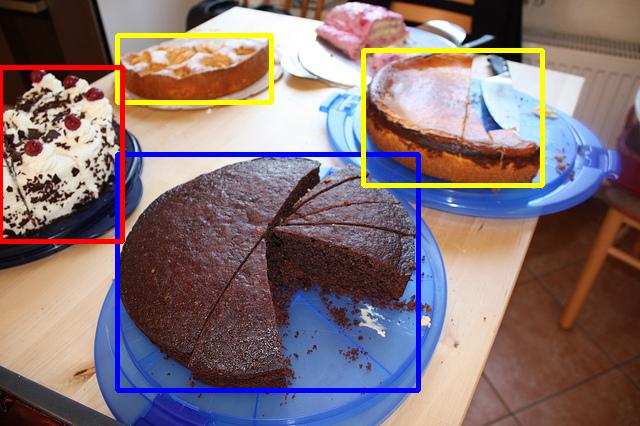}
    \end{center}\\
    \midrule
    \textbf{Text}: \textbf{elephant} on the left behind tree.\\
    \begin{center}
        \includegraphics[width=0.75\linewidth]{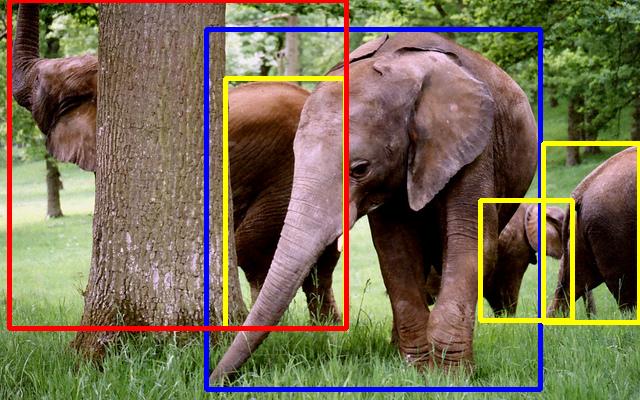}
    \end{center}\\
    \bottomrule
    \end{tabularx}
    \caption{Qualitative examples of selected regions from the RefCOCO validation set. We bold the words used for object detection. 
    We show the top prediction from \llavasmall{} in \textcolor{blue}{blue}, 
    the prediction from our method, 
    \llavasmall{} + \METHOD{} in \textcolor{red}{red}, and all remaining proposed regions from GroundingDINO in \textcolor{yellow}{yellow}.
    }
    \label{fig:qualitative_examples_bbox_refcoco}
\end{figure}

\section{Qualitative Examples}
\label{sec:qualitative_examples}

In \cref{fig:qualitative_examples_1}, \cref{fig:qualitative_examples_2}, \cref{fig:qualitative_examples_3}, and \cref{fig:qualitative_examples_4},
we include qualitative examples from ViP-Bench, where we show that \METHOD{} can direct the model to generate the correct answer to the given question. In \cref{fig:qualitative_examples_bbox_sugarcrepe}, we include examples of the correct text and its associated regions.
In \cref{fig:qualitative_examples_bbox_refcoco},
we include RefCOCO examples from the validation data with the bounding box proposals generated by GroundingDINO.

\end{document}